\documentclass[journal,11pt,draftclsnofoot,onecolumn]{IEEEtran}
\usepackage{setspace}
\usepackage{amssymb}
\usepackage{amsmath}
\usepackage{graphicx}
\usepackage{booktabs}
\usepackage{cite}
\usepackage{url}
\usepackage{enumerate}
\usepackage{tikz}
\usepackage[ruled,vlined]{algorithm2e}
\usetikzlibrary{shapes,decorations}
\usepgflibrary{shapes.geometric}
\usepackage{multirow}
\newcommand{\Alg}{\mathsf{Alg}}
\newcommand{\MinDegree}{\mathsf{MinDegree}}
\newcommand{\GreedyDegree}{\mathsf{GreedyDegree}}
\newcommand{\GreedyFillin}{\mathsf{GreedyFillin}}
\usepackage{comment}
\usepackage[usenames,dvipsnames]{pstricks}
\usepackage{epsfig}
\usepackage{subfigure}
\usepackage{enumerate}
\usepackage{bm}

\newtheorem{definition}{Definition}
\newtheorem{remark}{Remark}

\newtheorem{proposition}{Proposition}

\allowdisplaybreaks

\long\def\symbolfootnote[#1]#2{\begingroup%
 \def\thefootnote{\fnsymbol{footnote}}\footnote[#1]{#2}\endgroup}

\usepackage{cite}

\allowdisplaybreaks

\newcommand{\myfontsize}{\fontsize{9}{11}\selectfont}

\begin{document}

\def\Rd {{\mathbb {R}}^d}
\def\Rn {{\mathbb {R}}^N}
\def\Rm {{\mathbb {R}}^M}
\def\Nr {{\mathbb {N}}}
\def\Rmt {{\mathbb {R}}^{2M}}
\def\R {{\mathbb {R}}}
\def\Z {{\mathbb {Z}}}
\def\C {{\cal {C}}}
\def\Xb {{\mathbf{X}}}
\def\xb {{\mathbf{x}}}
\def\yb {{\mathbf{y}}}
\def\zb {{\mathbf{z}}}
\def\Mc {{\cal M}}
\def\Vc {{\cal V}}
\def\Uc {{\cal U}}
\def\eq#1{\begin{equation}{#1}\end{equation}}

\title{Finding Non-overlapping Clusters for Generalized Inference Over Graphical Models}
\author{Divyanshu Vats and Jos\'{e} M. F. Moura
\thanks{Divyanshu Vats is with the Institute for Mathematics and its Applications, University of Minnesota - Twin Cities, Minneapolis, MN, 55455, USA (email: dvats@ima.umn.edu)}
\thanks{Jos\'{e} M. F. Moura is with the Department of Electrical and Computer Engineering, Carnegie Mellon University, Pittsburgh, PA, 15213, USA (email: moura@ece.cmu.edu, ph: (412)-268-6341, fax: (412)-268-3980).}
}
\date{}

\maketitle

\begin{abstract}
Graphical models use graphs to compactly capture stochastic dependencies amongst a collection of random variables.  Inference over graphical models corresponds to finding marginal probability distributions given joint probability distributions.  In general, this is computationally intractable, which has led to a quest for finding efficient approximate inference algorithms.  We propose a framework for generalized inference over graphical models that can be used as a wrapper for improving the estimates of approximate inference algorithms. Instead of applying an inference algorithm to the original graph, we apply the inference algorithm to a block-graph, defined as a graph in which the nodes are non-overlapping clusters of nodes from the original graph. This results in marginal estimates of a cluster of nodes, which we further marginalize to get the marginal estimates of each node.  Our proposed block-graph construction algorithm is simple, efficient, and motivated by the observation that approximate inference is more accurate on graphs with longer cycles.  We present extensive numerical simulations that illustrate our block-graph framework with a variety  of inference algorithms (e.g., those in the libDAI software package).  These simulations show the improvements provided by our framework.
\end{abstract}


\pagebreak

\section{Introduction}

A graphical model is a probability distribution defined on a graph such that each node represents a random variable (or multiple random variables), and edges in the graph represent conditional independencies\footnote{In a graphical model over a graph $G$, for each edge $(i,j) \notin G$, $X_i$ is conditionally independent of $X_j$ given $X_{V \backslash \{i,j\}}$, where $V$ indexes all the nodes in the graph.  We can also say that for each $(i,j) \in G$, $X_i$ is dependent on $X_j$ given $X_S$, for all $S$ such that $S \subseteq V \backslash \{i,j\}$.}.  The underlying graph structure in a graphical model leads to a factorization of the joint probability distribution.  Graphical models are used in many applications such as sensor networks, image processing, computer vision, bioinformatics, speech processing, social network analysis, and ecology \cite{Reu2005,WainwrightJordan2008,KollerFriedman2009}, to name a few.  Inference over graphical models corresponds to finding the marginal distribution $p_s(x_s)$ for each random variable given the joint probability distribution $p(\xb)$.
It is well known that inference over graphical models is computationally tractable for only a small class of graphical models (graphs with low treewidth \cite{LauritzenSpiegelhalter1988}), which has led to much work to derive efficient approximate inference algorithms.

\subsection{Summary of Contributions}

\begin{figure}
\begin{center}
\scalebox{0.7} 
{
\begin{pspicture}(0,-2.2075)(15.879687,2.1675)
\definecolor{color38b}{rgb}{0.8,0.8,1.0}
\psframe[linewidth=0.04,dimen=outer,fillstyle=solid,fillcolor=color38b](9.061875,2.1675)(6.901875,1.1675)
\usefont{T1}{ptm}{m}{n}
\rput(7.96625,1.7025){\large $\Alg$}
\usefont{T1}{ptm}{m}{n}
\rput(4.8,1.6825){\large $(p(\xb),G)$}
\psline[linewidth=0.08cm,arrowsize=0.05291667cm 3.0,arrowlength=1.4,arrowinset=0.4]{->}(5.661875,1.6475)(6.641875,1.6475)
\psline[linewidth=0.08cm,arrowsize=0.05291667cm 3.0,arrowlength=1.4,arrowinset=0.4]{->}(9.321875,1.6675)(10.301875,1.6675)
\usefont{T1}{ptm}{m}{n}
\rput(11.090313,1.7025){\large $\{\widehat{p}_s(x_s)\}$}
\psframe[linewidth=0.04,dimen=outer,fillstyle=solid,fillcolor=color38b](9.061875,0.0675)(6.901875,-0.9325)
\usefont{T1}{ptm}{m}{n}
\rput(7.96625,-0.3975){\large $\Alg$}
\psline[linewidth=0.08cm,arrowsize=0.05291667cm 3.0,arrowlength=1.4,arrowinset=0.4]{->}(1.121875,-0.4325)(2.101875,-0.4325)
\usefont{T1}{ptm}{m}{n}
\rput(9.950313,-0.0375){$\{\widehat{p}_{\Vc_i}(x_{\Vc_i})\}$}
\psframe[linewidth=0.04,dimen=outer,fillstyle=solid,fillcolor=color38b](5.161875,0.0675)(2.361875,-0.9325)
\usefont{T1}{ptm}{m}{n}
\rput(3.776875,-0.4175){\large Block-Graph}
\usefont{T1}{ptm}{m}{n}
\rput(5.947031,-0.0975){$(p(\xb),{\cal G})$}
\psline[linewidth=0.08cm,arrowsize=0.05291667cm 3.0,arrowlength=1.4,arrowinset=0.4]{->}(5.421875,-0.4525)(6.761875,-0.4525)
\psframe[linewidth=0.04,dimen=outer,fillstyle=solid,fillcolor=color38b](13.141875,0.0475)(10.981875,-0.9525)
\usefont{T1}{ptm}{m}{n}
\rput(12.086094,-0.4375){\large Process}
\psline[linewidth=0.08cm,arrowsize=0.05291667cm 3.0,arrowlength=1.4,arrowinset=0.4]{->}(13.481875,-0.4125)(14.461875,-0.4125)
\usefont{T1}{ptm}{m}{n}
\rput(0.3,-0.4175){\large $(p(\xb),G)$}
\psline[linewidth=0.08cm,arrowsize=0.05291667cm 3.0,arrowlength=1.4,arrowinset=0.4]{->}(9.381875,-0.4525)(10.721875,-0.4525)
\usefont{T1}{ptm}{m}{n}
\rput(15.190312,-0.3775){\large $\{\widehat{p}_s(x_s)\}$}
\psframe[linewidth=0.04,linestyle=dashed,dash=0.16cm 0.16cm,dimen=outer](13.341875,0.6475)(2.181875,-1.4925)
\usefont{T1}{ptm}{m}{n}
\rput(7.9101562,-1.9375){\large Generalized Alg}
\end{pspicture} 
}
\end{center}
\caption{$\Alg$ is an inference algorithm that estimates marginal distributions given a graphical model.  We propose a framework that generalizes $\Alg$ using block-graphs to improve the accuracy of the marginal estimates.}
\label{fig:generalizedAlg}
\end{figure}
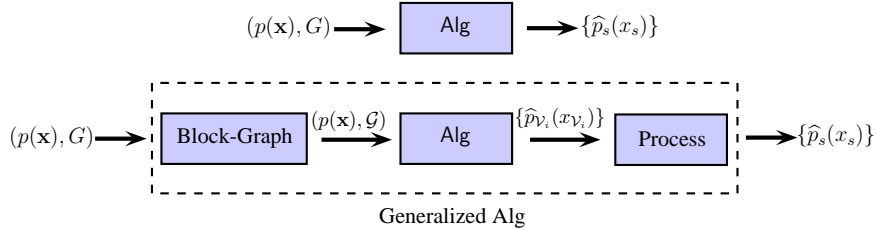

Our main contribution in this paper is a framework that can be used as a wrapper for improving the accuracy of approximate inference algorithms.  Instead of applying an inference algorithm to the original graph, we apply the inference algorithm to a block-graph, 
defined as a graph in which the nodes are non-overlapping clusters of nodes from the original graph.  This results in marginal estimates of a cluster of nodes of the original graph, which we further marginalize to get the marginal estimates of each node.  Larger clusters, in general, lead to more accurate inference algorithms at the cost of increased computational complexity.  Fig.~\ref{fig:generalizedAlg} illustrates our proposed block-graph framework for \emph{generalized inference}.

The key component in our framework is to construct a block-graph.   It has been empirically observed that approximate inference is more accurate on graphs with longer cycles \cite{tfabre-isit00}.  This motivates our proposed block-graph construction algorithm where we first find non-overlapping clusters such that the graph over the clusters is a tree.  We refer to the resulting block-graph as a block-tree.  The block-tree construction algorithm runs in linear time by using two passes of breadth-first search over the graph.  The second step in our block-graph construction algorithm is to split large clusters\footnote{For discrete graphical models, the complexity of inference is exponential in the maximum cluster size, which is why using the block-tree directly for inference may not be computationally tractable if the maximum cluster size is large.} in the block-tree.  Using numerical simulations, we show how our proposed algorithm for splitting large clusters leads to superior inference algorithms when compared to an algorithm that randomly splits large clusters in the block-tree and an algorithm that uses graph partitioning to find non-overlapping clusters \cite{XingGMFPartition2004}.

As an example, consider applying our block-graph framework to belief propagation (BP) \cite{Pearl1988}, which finds the marginal distribution at each node by iteratively passing messages between nodes.  If the graph is a tree, BP computes the exact marginal distributions, however, for general graphs with cycles, BP only approximates the true marginal distribution.  Our framework for inference (see Fig.~\ref{fig:generalizedAlg}) generalizes BP so that message passing occurs between clusters of nodes, where the clusters are non-overlapping.  The estimates of the marginal distribution at each node can be computed by marginalizing the approximate marginal distribution of each cluster.  Our framework is not limited to BP and can be used as a wrapper for \emph{any} inference algorithm on graphical models as we show in Section~\ref{sec:approximateinference}.  Using numerical simulations, we show how our block-graph framework improves the marginal distribution estimates computed by current inference algorithms in the literature: BP \cite{Pearl1988}, conditioned belief propagation (CBP) \cite{Eaton09}, loop corrected belief propagation (LC) \cite{MontanariRizzo2005,MooijLC2007}, tree-structured expectation propagation (TreeEP), iterative join-graph propagation (IJGP) \cite{MateescuKGDIJGP2010}, and generalized belief propagation (GBP) \cite{YedidaGBP2001,HAK2003,YedidaTIT2005,Pelizzola2005}.

\subsection{Related Work}
There has been significant work in extending the BP algorithm of message passing between nodes to message passing between clusters.  It is known that the true marginal distributions of a graphical model minimize the Gibbs free energy \cite{JordanVariational1999}.  In \cite{YedidaGBP2001,YedidaTRBethe2001}, the authors show that the fixed points of the BP algorithm minimize the Bethe free energy, which is an approximation to the Gibbs free energy.  This motivated the generalized belief propagation (GBP) algorithm that minimizes the Kikuchi free energy \cite{Kikuchi1951}, a better approximation to the Gibbs free energy.  In GBP, message passing is between clusters of nodes that are overlapping.  A more general approach to GBP is proposed in \cite{YedidaTIT2005} using region graphs and in \cite{Pelizzola2005} using the cluster variation method (CVM).  References \cite{WellingUAI2004,WelMinTeh2005} propose some guidelines for choosing clusters in GBP.  Reference \cite{MateescuKGDIJGP2010} proposes a framework for GBP, called iterative join-graph propagation (IJGP), that first constructs a junction tree, a tree-structured representation of a graph with overlapping clusters, and then splits larger clusters in the junction tree to perform message passing over a set of overlapping clusters.  

In the original paper describing GBP \cite{YedidaTRBethe2001}, the authors give an example of how non-overlapping clusters can be used for GBP, since, when applying the block-graph framework to BP, the resulting inference algorithm (B$m$-BP, where $m$ is the cluster size) becomes a class of GBP algorithms where the set of overlapping clusters corresponds to cliques in the block-graph.  Our numerical simulations identify cases where B$m$-BP leads to superior marginal estimates when compared to a GBP algorithm that uses overlapping clusters.  Moreover, since our framework can be applied to any inference algorithm, we show that the marginal estimates computed by GBP based algorithms can be improved by applying the GBP based algorithm to a block-graph.  Our block-graph framework is not limited to generalizing BP and we show this in our numerical simulations where we generalize conditioned belief propagation (CBP) and loop corrected belief propagation (LC).  Both these algorithms have been shown to empirically perform better than GBP for certain graphical models \cite{MooijLC2007,Eaton09}.  In \cite{XingGMF2003,XingGMFPartition2004}, the authors propose using graph partitioning algorithms for finding non-overlapping clusters for generalizing the mean field algorithm for inference \cite{JordanVariational1999}.  Our numerical results show that our algorithm for finding non-overlapping clusters leads to superior marginal estimates.

We note that our work differs from some other works on studying graphical models defined over graphs with non-overlapping clusters.  For example, \cite{MarlinMurphy2009} consider the problem of learning a Gaussian graphical model defined over some block-graph.  Similar efforts have been made in \cite{JalaliRaviSang2011} for discrete valued graphical models.  In \cite{MalioutovThesis}, the author analyzes properties of a graphical model defined on a block-tree.  In all of the above works, the underlying graphical model is \emph{assumed} to be block-structured.  In our work, we assume a graphical model defined on an \emph{arbitrary} graph and then find a representation of the graphical model on a block-graph to enable more accurate inference algorithms.

This paper is motivated by our earlier work in studying tree structures for Markov random fields (MRFs) indexed over continuous indices \cite{Levy1956}.  In \cite{VatsMoura2009j}, we have shown that a natural tree-like representation for such MRFs exists over non-overlapping \emph{hypersurfaces} within the continuous index set.  Using this representation, we derived extensions of the Kalman-Bucy filter \cite{KalmanBucy1961} and the Rauch-Tung-Striebel smoother \cite{RauchTungStriebel1965} to Gaussian MRFs indexed over continuous indices.

\subsection{Paper Organization}
Section~\ref{sec:backgroud} reviews graphical models and the inference problem.  Section~\ref{sec:bt} outlines our proposed algorithm for constructing block-trees, a tree-structured graph over non-overlapping clusters.  Section~\ref{sec:blockgraph} presents our algorithm for splitting larger clusters in a block-tree  to construct a block-graph.  Section~\ref{sec:approximateinference} outlines our block-graph framework for generalizing inference algorithms.  Section~\ref{sec:simulations} presents extensive numerical results evaluating our framework on various inference algorithms.  Section~\ref{sec:summary} summarizes the paper and outlines some future research directions.

\section{Background: Graphical Models and Inference}
\label{sec:backgroud}

\begin{figure}
\begin{center}
\scalebox{1.0} 
{
\begin{pspicture}(0,-1.3503125)(5.970924,2.3503125)
\definecolor{color809b}{rgb}{0.6,0.6,0.6}
\pscircle[linewidth=0.04,dimen=outer](1.2724775,0.7065626){0.28}
\usefont{T1}{ptm}{m}{n}
\rput(1.2466962,0.69656265){\small $x_3$}
\pscircle[linewidth=0.04,dimen=outer](2.5924776,1.0665625){0.28}
\usefont{T1}{ptm}{m}{n}
\rput(2.5666962,1.0565625){\small $x_9$}
\pscircle[linewidth=0.04,dimen=outer,fillstyle=solid,fillcolor=color809b](0.71247756,-0.45343745){0.28}
\usefont{T1}{ptm}{m}{n}
\rput(0.68669623,-0.44343746){\small $x_1$}
\pscircle[linewidth=0.04,dimen=outer,fillstyle=solid,fillcolor=color809b](2.8524776,0.00656255){0.28}
\usefont{T1}{ptm}{m}{n}
\rput(2.8266962,0.01656255){\small $x_4$}
\pscircle[linewidth=0.04,dimen=outer](1.9324777,-0.89343745){0.28}
\usefont{T1}{ptm}{m}{n}
\rput(1.9066962,-0.88343745){\small $x_2$}
\pscircle[linewidth=0.04,dimen=outer,fillstyle=solid,fillcolor=color809b](3.3124774,-1.1534375){0.28}
\usefont{T1}{ptm}{m}{n}
\rput(3.2866964,-1.1634375){\small $x_5$}
\pscircle[linewidth=0.04,dimen=outer](4.372478,-0.6734375){0.28}
\usefont{T1}{ptm}{m}{n}
\rput(4.346696,-0.68343747){\small $x_6$}
\pscircle[linewidth=0.04,dimen=outer](4.0924773,0.36656255){0.28}
\usefont{T1}{ptm}{m}{n}
\rput(4.066696,0.37656254){\small $x_7$}
\pscircle[linewidth=0.04,dimen=outer](3.892478,1.2865626){0.28}
\usefont{T1}{ptm}{m}{n}
\rput(3.866696,1.2965626){\small $x_8$}
\rput{-25.676483}(0.041561563,0.43831116){\psellipse[linewidth=0.04,linestyle=dotted,dotsep=0.16cm,dimen=outer](0.98242635,0.12797038)(0.49574608,1.2362015)}
\rput{17.648941}(0.11927368,-0.89626944){\psellipse[linewidth=0.04,linestyle=dotted,dotsep=0.16cm,dimen=outer](2.946256,-0.06398947)(0.5213262,1.7151822)}
\rput{13.837411}(0.19320941,-0.9763108){\psellipse[linewidth=0.04,linestyle=dotted,dotsep=0.16cm,dimen=outer](4.119491,0.30796355)(0.50680095,1.5024654)}
\usefont{T1}{ptm}{m}{n}
\rput(0.94560254,1.6365626){A}
\usefont{T1}{ptm}{m}{n}
\rput(2.4668524,1.9965625){B}
\usefont{T1}{ptm}{m}{n}
\rput(3.8668525,2.1765625){C}
\psline[linewidth=0.04cm](0.77092475,-0.18968745)(1.1709248,0.47031254)
\psline[linewidth=0.04cm](0.8709248,-0.6296874)(1.6709248,-0.86968744)
\psline[linewidth=0.04cm](1.4509248,0.89031255)(2.3309247,1.0503125)
\psline[linewidth=0.04cm](2.0709248,-0.66968745)(2.6909246,-0.16968745)
\psline[linewidth=0.04cm](2.6309247,0.81031257)(2.7709248,0.27031255)
\psline[linewidth=0.04cm](2.1509247,-1.0096874)(3.0509248,-1.1696875)
\psline[linewidth=0.04cm](3.0709248,-0.10968745)(4.1309247,-0.58968747)
\psline[linewidth=0.04cm](3.5309248,-1.0496875)(4.1509247,-0.80968744)
\psline[linewidth=0.04cm](4.1509247,0.13031255)(4.330925,-0.40968746)
\psline[linewidth=0.04cm](3.9509249,1.0503125)(4.050925,0.63031256)
\psline[linewidth=0.04cm](2.8309247,1.1503125)(3.6309247,1.2303126)
\end{pspicture} 
}
\caption{An example of a graphical model.  The global Markov property states that $x_A$ is independent of $x_C$ given $x_B$ since all paths from $A$ to $C$ pass through $B$.}
\label{fig:example_gm}
\end{center}
\end{figure}
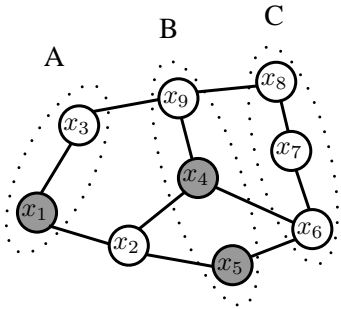

A graphical model is defined using a graph $G = (V,E)$, where the nodes $V = \{1,2,\ldots,p\}$ index a collection of random variables $\xb = \{x_s \in \Omega^d: s \in V\}$ and the edges $E \subseteq V \times V$ encode statistical independencies \cite{Lauritzen1996,KollerFriedman2009}.  The set of edges can be directed, undirected, or both.  Since directed graphical models, also known as Bayesian networks, can be mapped to undirected graphical models by moralizing the graph, in this paper, we only consider undirected graphical models, also known as Markov random fields or Markov networks.

The edges in a graphical model imply Markov properties about the collection of random variables.  The \emph{local Markov property} states that $x_s$ is independent of $\{x_r: r \in V \backslash \{{\cal N}(s)\cup s\}\}$ given $x_{{\cal N}(s)}$, where ${\cal N}(s)$ is the set of neighbors of $s$.  For example, in Fig.~\ref{fig:example_gm}, $x_2$ is independent of $\{x_3,x_6,x_7,x_8,x_9\}$ given $\{x_1,x_4,x_5\}$.  The \emph{global Markov  property}, which is equivalent to the local Markov property for non-degenerate probability distributions, states that, for a collection of disjoint nodes $A$, $B$, and $C$, if $B$ separates $A$ and $C$, $x_A$ is independent of $x_C$ given $x_B$.  An example of the sets $A$, $B$, and $C$ is shown in Fig.~\ref{fig:example_gm}. From the \emph{Hammersley-Clifford} theorem \cite{Besag1974}, the Markov property leads to a factorization of the joint probability distribution over cliques (fully connected subsets of nodes) in the graph,
\begin{equation}
p(x_1,x_2,\ldots,x_p) = \frac{1}{Z} \prod_{C \in {\cal C}} \psi_C(x_C) \,,
\label{eq:dist_mrf}
\end{equation}
where ${\cal C}$ is the set of all cliques in the graph $G = (V,E)$, $\psi_C(x_C) > 0$ are potential functions defined over cliques, and $Z$ is the partition function, a normalization constant.

Inference in graphical models corresponds to finding marginal distributions, say $p_s(x_s)$, given the probability distribution $p(\xb)$ for $\xb = \{x_1,\ldots,x_p\}$.  This problem is of extreme importance in many domains.  A classical example is in estimation when we are given noisy observations $\yb$ of $\xb$ and we want to estimate the underlying random vector.  To find the minimum mean square error (mmse) estimate, we need to marginalize the conditional probability distribution $p(\xb | \yb)$ to find the marginals $p_s(x_s|\yb)$.  An algorithm for marginalizing $p(\xb)$ can be used for marginalizing $p(\xb | \yb)$.  In general, exact inference is computationally intractable, however, there has been significant progress in deriving efficient approximate inference algorithms.  The main contribution in this paper is the block-graph framework for generalizing inference algorithms (see Fig.~\ref{fig:generalizedAlg}) so that the performance of approximate inference algorithms can be improved.


\section{Block-Trees: Finding Trees Over Non-overlapping Clusters}
\label{sec:bt}

Section~\ref{subsec:mainalgorithmblocktrees} outlines our algorithm for constructing block-trees.  Section~\ref{sec:optimalbt} defines the notion of optimal block-trees by using connections between block-trees and junction trees.  Section~\ref{sec:greedybt} outlines greedy algorithms for finding optimal block-trees.

\subsection{Main Algorithm}
\label{subsec:mainalgorithmblocktrees}

\begin{definition}[Block-Graph and Block-Tree]
\label{def:blocktree}
For a graph $G = (V,E)$, a \emph{block-graph} ${\cal G} = ({\cal V},{\cal E})$ is a graph over clusters of nodes in $V$ such that each node in $V$ is associated with only one cluster in $\Vc$.  In other words, the clusters in ${\cal V}$ are non-overlapping.  If the edge set ${\cal E} \subseteq V \times V$ is tree-structured, we call the block-graph a \emph{block-tree}.
\end{definition}

Algorithm~\ref{alg:blocktree} outlines our construction of a block-tree ${\cal G}$ given an arbitrary graph $G = (V,E)$.  Without loss in generality, we assume that $G$ is connected, i.e., there exists a path between all non adjacent nodes.  The original graph $G$ and a set of nodes $V_1 \subset V$ are the input.  The output is the block-tree ${\cal G}$.  We refer to $V_1$ as the \emph{root cluster}.  The algorithm first finds an initial set of clusters and then splits these clusters to find the final block-tree.  We explain the steps of the algorithm.

\noindent
\textbf{Forward step}: Find clusters $V_1,V_2,\ldots,V_r$ using breadth-first search (BFS) so that $V_2 = {\cal N}(V_1),V_3 = {\cal N}(V_2) \backslash \{V_1 \cup V_2\},\ldots,V_r = {\cal N}(V_r) \backslash \{V_{r-2} \cup V_{r-1}\}$ .  These clusters serve as initial clusters for the block-tree.  During the BFS step, split each cluster $V_k$ into its connected components $\{V_k^1,\ldots,V_k^{m_k}\}$ using the subgraph $G(V_k)$, which denotes the graph only over the nodes in $V_k$ (Line 2).

\noindent
\textbf{Backwards step}:  We now merge the clusters in each $V_k$ to find the final block-tree.  The key intuition in this step is that each cluster $V_k$ should be connected to a single cluster in $V_{k-1}$.  If this is not the case, we merge clusters in $V_{k-1}$ accordingly.  Starting at $V_r = \{V_r^1,V_r^2,\ldots,V_r^{m_r}\}$, for each $V_r^j, j = 1,\ldots,m_r$, find all clusters $C(V_r^j)$ in $V_{r-1}$ that are connected to $V_r^j$ (Line 6).  Combine all clusters in $C(V_r^j)$ into a single cluster and update the clusters in $V_{r-1}^j$ accordingly.  Repeat the above steps for all the clusters in $V_{r-1},V_{r-2},\ldots,V_3$.

\begin{algorithm}
\label{alg:blocktree}
\DontPrintSemicolon
\KwData{A graph $G = (V,E)$ and a set of nodes $V_1$.}
\KwResult{A block-tree ${\cal G} = ({\cal C},{\cal E})$}
\caption{Constructing Block-Trees: BlockTree($G$,$V_1$)}
\nl Find successive neighbors of $V_1$ to construct a sequence of $r$ clusters $V_1,V_2,\ldots,V_r$ such that $V_2 = {\cal N}(V_1)$, $V_3 = {\cal N}(V_2) \backslash \{V_1 \cup V_2\}, \ldots, V_r = {\cal N}(V_r) \backslash \{V_{r-2} \cup V_{r-1}\}$.  \;
\nl $\{V_k^1,\ldots,V_k^{m_k}\} \gets $ Find $m_k$ connected components of $V_k$ using subgraph $G(V_k)$. \;
\nl \For{$i = r,r-1,\ldots 3$}{
\nl \For{$j = 1,2,\ldots m_{i}$}{
\nl $C(V_{i}^j) \gets {\cal N}(V_i^j) \cap V_{i-1}$ ; All nodes in $V_{i-1}$ connected to $V_i^j$.  \;
\nl Combine $C(V_{i}^j)$ into one cluster and update $V_{i-1}$.
}
}
\nl ${\cal V} \gets \bigcup_{k = 1}^{r} \{V_k^1,V_k^2,\ldots,V_k^{m_k} \}$ \;
\nl ${\cal E} \gets \text{edges between all the clusters in ${\cal V}$}$
\end{algorithm}

The first part of Algorithm~\ref{alg:blocktree} finds successive non-overlapping neighbors of the root cluster.  This leads to an initial estimate of the block-tree graph.  In the backwards step, we split clusters to form a block-tree.  We illustrate Algorithm~\ref{alg:blocktree} with examples.

\begin{figure}
\begin{center}
\subfigure[]{
\begin{tikzpicture}[scale=0.6]
\tikzstyle{every node}=[draw,shape=circle,scale=0.6,fill=blue!20];
\path (0,5) node (x1) {$1$};
\path (-1,4) node (x4) {$4$};
\path (1,4) node (x2) {$2$};
\path (-2,3) node (x7) {$7$};
\path (0,3) node (x5) {$5$};
\path (2,3) node (x3) {$3$};
\path (-1,2) node (x8) {$8$};
\path (1,2) node (x6) {$6$};
\path (0,1) node (x9) {$9$};
\draw (x1) -- (x2) -- (x3) -- (x6) -- (x9) -- (x8) -- (x7);
\draw (x7) -- (x4) -- (x1) (x4) -- (x5) -- (x6);
\draw (x2) -- (x5) -- (x8);
\tikzstyle{every node}=[scale=1.0];
\draw (-0.4,5.4) -- (0.4,5.4) -- (0.4,4.6) -- (-0.4,4.6) -- cycle;
\node (tt) at (0.85,5) {$V_1$};
\draw (-1.4,4.4) -- (1.4,4.4) -- (1.4,3.6) -- (-1.4,3.6) -- cycle;
\node (tt1) at (1.9,4) {$V_2$};
\draw (-2.4,3.4) -- (2.4,3.4) -- (2.4,2.6) -- (-2.4,2.6) -- cycle;
\node (tt1) at (2.9,3) {$V_3$};
\draw (-1.4,2.4) -- (1.4,2.4) -- (1.4,1.6) -- (-1.4,1.6) -- cycle;
\node (tt1) at (1.9,2) {$V_4$};
\draw (-0.4,1.4) -- (0.4,1.4) -- (0.4,0.6) -- (-0.4,0.6) -- cycle;
\node (tt) at (0.85,1) {$V_5$};
\draw [ultra thick] [->] (-2.8,5) -- (-2.8,1);
\end{tikzpicture}
} \quad
\subfigure[]{
\begin{tikzpicture}[scale=0.6]
\tikzstyle{every node}=[draw,shape=circle,scale=0.6,fill=blue!20];
\path (0,5) node (x1) {$1$};
\path (-1,4) node (x4) {$4$};
\path (1,4) node (x2) {$2$};
\path (-2,3) node (x7) {$7$};
\path (0,3) node (x5) {$5$};
\path (2,3) node (x3) {$3$};
\path (-1,2) node (x8) {$8$};
\path (1,2) node (x6) {$6$};
\path (0,1) node (x9) {$9$};
\draw (x1) -- (x2) -- (x3) -- (x6) -- (x9) -- (x8) -- (x7);
\draw (x7) -- (x4) -- (x1) (x4) -- (x5) -- (x6);
\draw (x2) -- (x5) -- (x8);
\tikzstyle{every node}=[scale=1.0];
\draw (-0.4,5.4) -- (0.4,5.4) -- (0.4,4.6) -- (-0.4,4.6) -- cycle;
\draw (-1.4,4.4) -- (1.4,4.4) -- (1.4,3.6) -- (-1.4,3.6) -- cycle;
\draw (-2.4,3.4) -- (2.4,3.4) -- (2.4,2.6) -- (-2.4,2.6) -- cycle;
\draw (-1.4,2.4) -- (1.4,2.4) -- (1.4,1.6) -- (-1.4,1.6) -- cycle;
\draw (-0.4,1.4) -- (0.4,1.4) -- (0.4,0.6) -- (-0.4,0.6) -- cycle;
\draw [ultra thick] [->] (3.3,1) -- (3.3,5);
\end{tikzpicture}
} \qquad
\subfigure[]{
\begin{tikzpicture}[scale=0.6]
\tikzstyle{every node}=[draw,scale=0.6,fill=blue!20];
\path (0,4) node[circle] (c1) {$1$};
\path (0,3) node[ellipse] (c2) {$2$ $4$};
\path (0,2) node[ellipse] (c3) {$3$ $5$ $7$};
\path (0,1) node[ellipse] (c4) {$6$ $8$};
\path (0,0) node[circle] (c5) {$9$};
\tikzstyle{every node}=[scale=0.7];
\draw (c1) -- (c2) -- (c3) -- (c4) -- (c5);
\end{tikzpicture}
}
\caption{{\small{(a) Original estimates of the clusters in a grid graph when running the forward pass of Algorithm~\ref{alg:blocktree}. (b) The final clusters after running the backwards pass of Algorithm~\ref{alg:blocktree}. (c) Final block-tree. }}}
\label{fig:example-grid-bt}
\end{center}
\end{figure}
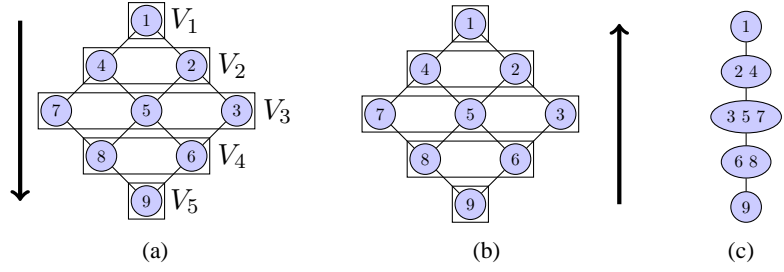

\noindent
\textbf{Example:} Consider the grid graph of Fig.~\ref{fig:example-grid-bt}(a).  Choosing $V_1 = \{1\}$, we get the initial estimates of the clusters as shown in Fig.~\ref{fig:example-grid-bt}(a).  Running the backwards step to identify the final clusters (see Fig.~\ref{fig:example-grid-bt}(b)), we get the block-tree in Fig.~\ref{fig:example-grid-bt}(c).

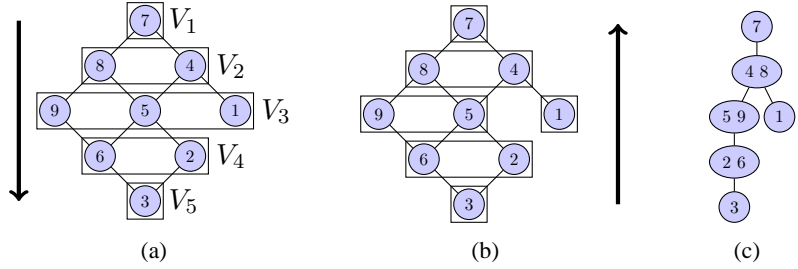
\begin{figure}
\begin{center}
\subfigure[]{
\begin{tikzpicture}[scale=0.6]
\tikzstyle{every node}=[draw,shape=circle,scale=0.6,fill=blue!20];
\path (0,5) node (x7) {$7$};
\path (-1,4) node (x8) {$8$};
\path (1,4) node (x4) {$4$};
\path (-2,3) node (x9) {$9$};
\path (0,3) node (x5) {$5$};
\path (2,3) node (x1) {$1$};
\path (-1,2) node (x6) {$6$};
\path (1,2) node (x2) {$2$};
\path (0,1) node (x3) {$3$};
\draw (x2) -- (x3) -- (x6) -- (x9) -- (x8) -- (x7);
\draw (x7) -- (x4) -- (x1) (x4) -- (x5) -- (x6);
\draw (x2) -- (x5) -- (x8);
\tikzstyle{every node}=[scale=1.0];
\draw (-0.4,5.4) -- (0.4,5.4) -- (0.4,4.6) -- (-0.4,4.6) -- cycle;
\node (tt) at (0.85,5) {$V_1$};
\draw (-1.4,4.4) -- (1.4,4.4) -- (1.4,3.6) -- (-1.4,3.6) -- cycle;
\node (tt1) at (1.9,4) {$V_2$};
\draw (-2.4,3.4) -- (2.4,3.4) -- (2.4,2.6) -- (-2.4,2.6) -- cycle;
\node (tt1) at (2.9,3) {$V_3$};
\draw (-1.4,2.4) -- (1.4,2.4) -- (1.4,1.6) -- (-1.4,1.6) -- cycle;
\node (tt1) at (1.9,2) {$V_4$};
\draw (-0.4,1.4) -- (0.4,1.4) -- (0.4,0.6) -- (-0.4,0.6) -- cycle;
\node (tt) at (0.85,1) {$V_5$};
\draw [ultra thick] [->] (-2.8,5) -- (-2.8,1);
\end{tikzpicture}
} \quad
\subfigure[]{
\begin{tikzpicture}[scale=0.6]
\tikzstyle{every node}=[draw,shape=circle,scale=0.6,fill=blue!20];
\path (0,5) node (x7) {$7$};
\path (-1,4) node (x8) {$8$};
\path (1,4) node (x4) {$4$};
\path (-2,3) node (x9) {$9$};
\path (0,3) node (x5) {$5$};
\path (2,3) node (x1) {$1$};
\path (-1,2) node (x6) {$6$};
\path (1,2) node (x2) {$2$};
\path (0,1) node (x3) {$3$};
\draw (x2) -- (x3) -- (x6) -- (x9) -- (x8) -- (x7);
\draw (x7) -- (x4) -- (x1) (x4) -- (x5) -- (x6);
\draw (x2) -- (x5) -- (x8);
\tikzstyle{every node}=[scale=1.0];
\draw (-0.4,5.4) -- (0.4,5.4) -- (0.4,4.6) -- (-0.4,4.6) -- cycle;
\draw (-1.4,4.4) -- (1.4,4.4) -- (1.4,3.6) -- (-1.4,3.6) -- cycle;
\draw (-2.4,3.4) -- (0.4,3.4) -- (0.4,2.6) -- (-2.4,2.6) -- cycle;
\draw (1.6,3.4) -- (2.4,3.4) -- (2.4,2.6) -- (1.6,2.6) -- cycle;
\draw (-1.4,2.4) -- (1.4,2.4) -- (1.4,1.6) -- (-1.4,1.6) -- cycle;
\draw (-0.4,1.4) -- (0.4,1.4) -- (0.4,0.6) -- (-0.4,0.6) -- cycle;
\draw [ultra thick] [->] (3.3,1) -- (3.3,5);
\end{tikzpicture}
} \qquad
\subfigure[]{
\begin{tikzpicture}[scale=0.6]
\tikzstyle{every node}=[draw,scale=0.6,fill=blue!20];
\path (0,4) node[circle] (c1) {$7$};
\path (0,3) node[ellipse] (c2) {$4$ $8$};
\path (-0.5,2) node[ellipse] (c3) {$5$ $9$};
\path (0.5,2) node[circle] (c4) {$1$};
\path (-0.5,1) node[ellipse] (c5) {$2$ $6$};
\path (-0.5,0) node[circle] (c6) {$3$};
\tikzstyle{every node}=[scale=0.7];
\draw (c1) -- (c2) -- (c3) -- (c5) -- (c6);
\draw (c2) -- (c4);
\end{tikzpicture}
}
\caption{{\small{(a) Original estimates of the clusters in a partial grid when running the forward pass of Algorithm~\ref{alg:blocktree}. (b) The final clusters after running the backwards pass of Algorithm~\ref{alg:blocktree}. (c) Final block-tree. }}}
\label{fig:partialgrid-bt}
\end{center}
\end{figure}

\noindent
\textbf{Example:} In the previous example, the initial estimates of the clusters matched the final estimates and the final block-tree was a chain-structured graph.  We now consider an example where the final block-tree will in fact be tree-structured.  Consider the partial grid graph of Fig.~\ref{fig:partialgrid-bt}(a).  Choosing $V_1 = \{7\}$, we get the initial estimates of the clusters in Fig.~\ref{fig:partialgrid-bt}(a).  We now run the backwards step of the algorithm.  Since $V_5 = \{3\}$ is connected to $2$ and $6$, $C(V_5) = \{2,6\}$.  Thus, $\{2,6\}$ become a single cluster.  We now find neighbors of $\{2,6\}$ in $V_3 = \{9,5,1\}$.  It is clear that only $\{9,5\}$ are connected to $\{2,6\}$, so $\{9,5\}$ become a single cluster.  In this way, we have split $V_3$ into two clusters: $V_3^1 = \{9,5\}$ and $V_3^2 = \{1\}$.  Continuing the algorithm, we find the remaining clusters as shown in Fig.~\ref{fig:partialgrid-bt}(b).  The final block-tree is shown in Fig.~\ref{fig:partialgrid-bt}(c).

The following proposition characterizes the time complexity and correctness of Algorithm~\ref{alg:blocktree}.

\begin{proposition}
\label{prop:btcomplexity}
Algorithm~\ref{alg:blocktree} runs in time $O(|E|)$ and always outputs a block-tree.
\end{proposition}
\begin{IEEEproof}
Both the forward step and the backwards step involve a breadth first search, which has complexity $O(|E|)$.
Algorithm~\ref{alg:blocktree} always outputs a block-tree since each cluster in $V_k$ is only connected to a single cluster in $V_{k-1}$.
\end{IEEEproof}

Block-trees are closely related to junction trees.  In the next Section, we explore this connection to define optimal block-trees.



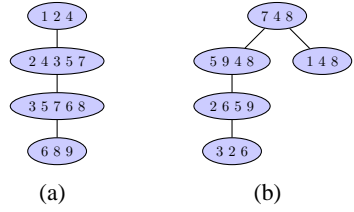
\begin{figure}
\begin{center}
\subfigure[]{
\begin{tikzpicture}[scale=0.6]
\tikzstyle{every node}=[draw,scale=0.5,fill=blue!20];
\path (0,4) node[ellipse] (c1) {$1$ $2$ $4$};
\path (0,3) node[ellipse] (c2) {$2$ $4$ $3$ $5$ $7$};
\path (0,2) node[ellipse] (c3) {$3$ $5$ $7$ $6$ $8$};
\path (0,1) node[ellipse] (c4) {$6$ $8$ $9$};
\tikzstyle{every node}=[scale=0.7];
\draw (c1) -- (c2) -- (c3) -- (c4);
\end{tikzpicture}
} \qquad
\subfigure[]{
\begin{tikzpicture}[scale=0.6]
 \tikzstyle{every node}=[draw,scale=0.5,fill=blue!20];
\path (0,4) node[ellipse] (c1) {$7$ $4$ $8$};
\path (1,3) node[ellipse] (c2) {$1$ $4$ $8$};
\path (-1,3) node[ellipse] (c3) {$5$ $9$ $4$ $8$};
\path (-1,2) node[ellipse] (c4) {$2$ $6$ $5$ $9$};
\path (-1,1) node[ellipse] (c5) {$3$ $2$ $6$};
\tikzstyle{every node}=[scale=0.7];
\draw (c2) -- (c1) -- (c3) -- (c4) -- (c5);
\end{tikzpicture}
}
\caption{{{(a) Junction tree for the block-tree in Fig.~\ref{fig:example-grid-bt}(c) (b) Junction tree for the block-tree in Fig.~\ref{fig:partialgrid-bt}(c)}}}
\label{fig:btTojt}
\end{center}
\end{figure}

\subsection{Optimal Block-Trees}
\label{sec:optimalbt}

Junction trees, also known as clique trees or join trees, are tree-structured representations of graphs using a set of overlapping clusters \cite{HalinTreeDecomp1976, Robertson1986}.  The \textit{width} of a graph is the maximum cardinality of a cluster in the junction tree minus one.  The \textit{treewidth} of a graph is the minimum width of a graph over all possible junction tree representations.  It is well known that several graph related problems that are computationally intractable in general can be solved efficiently when the graph has low treewidth.  For the problem of inference over graphical models, \cite{LauritzenSpiegelhalter1988} showed how junction trees can be used for exact inference over graphical models defined over graphs with small treewidth.

Given a block-tree ${\cal G} = ({\cal V},{\cal E})$, an equivalent junction tree representation can be easily computed by combining all clusters connected along edges into a single cluster.  For example, the junction tree representation for the block-trees in Fig.~\ref{fig:example-grid-bt}(c) and Fig.~\ref{fig:partialgrid-bt}(c) are given in Fig.~\ref{fig:btTojt}(a) and Fig.~\ref{fig:btTojt}(b), respectively.  Using this junction tree, we can derive exact inference algorithms for graphical models parameterized by block-trees.  

It is easy to see that the complexity of inference using block-trees will depend on the maximum sum of cluster sizes of adjacent clusters in the block-tree (since this will correspond to the width of the equivalent junction tree).  Thus, an \textit{optimal block-tree} can be defined as a block-tree ${\cal G} = ({\cal V},{\cal E})$ for which $ \max_{(i,j) \in {\cal E}} \left( |V_i| + |V_j| \right)$ is minimized.  From Algorithm~\ref{alg:blocktree}, the construction of the block-tree depends on the choice of the root cluster $V_1$.  Thus, finding an \emph{optimal block-tree} is equivalent to finding an optimal root cluster.  
This problem is {computationally intractable} since we need to search over all possible combinations of root clusters $V_1$.

\begin{figure}
\begin{center}
\subfigure[]{
\begin{tikzpicture}[scale=0.7]
\tikzstyle{every node}=[draw,shape=circle,scale=0.6,fill=blue!20];
\path (-0.5,3) node (x7) {$7$};
\path (0.5,3) node (x4) {$4$};
\path (-1,2) node (x8) {$8$};
\path (0,2) node (x5) {$5$};
\path (1,2) node (x1) {$1$};
\path (-1,1) node (x9) {$9$};
\path (0,1) node (x6) {$6$};
\path (1,1) node (x2) {$2$};
\path (0,0) node (x3) {$3$};
\draw (x2) -- (x3) -- (x6) -- (x9) -- (x8) -- (x7) -- (x4) -- (x5) -- (x2);
\draw (x6) -- (x5) -- (x8);
\draw (x4) -- (x1);
\tikzstyle{every node}=[scale=1.1];
\draw (-0.8,3.4) -- (0.8,3.4) -- (0.8,2.6) -- (-0.8,2.6) -- cycle;
\node (tt) at (1.2,2.9) {$V_1$};
\draw (-1.3,2.4) -- (1.3,2.4) -- (1.3,1.6) -- (-1.3,1.6) -- cycle;
\node (tt) at (1.7,1.9) {$V_2$};
\draw (-1.3,1.4) -- (1.3,1.4) -- (1.3,0.6) -- (-1.3,0.6) -- cycle;
\node (tt) at (1.7,0.9) {$V_3$};
\draw (-0.3,0.4) -- (0.3,0.4) -- (0.3,-0.4) -- (-0.3,-0.4) -- cycle;
\node (tt) at (0.7,0.0) {$V_4$};
\end{tikzpicture}
} \qquad
\subfigure[]{
\begin{tikzpicture}[scale=0.7]
\tikzstyle{every node}=[draw,shape=circle,scale=0.6,fill=blue!20];
\path (-0.5,3) node (x7) {$7$};
\path (0.5,3) node (x4) {$4$};
\path (-1,2) node (x8) {$8$};
\path (0,2) node (x5) {$5$};
\path (1,2) node (x1) {$1$};
\path (-1,1) node (x9) {$9$};
\path (0,1) node (x6) {$6$};
\path (1,1) node (x2) {$2$};
\path (0,0) node (x3) {$3$};
\draw (x2) -- (x3) -- (x6) -- (x9) -- (x8) -- (x7) -- (x4) -- (x5) -- (x2);
\draw (x6) -- (x5) -- (x8);
\draw (x4) -- (x1);
\tikzstyle{every node}=[scale=1.0];
\draw (-0.8,3.4) -- (0.8,3.4) -- (0.8,2.6) -- (-0.8,2.6) -- cycle;
\draw (-1.3,2.4) -- (0.3,2.4) -- (0.3,1.6) -- (-1.3,1.6) -- cycle;
\draw (0.7,2.4) -- (1.3,2.4) -- (1.3,1.6) -- (0.7,1.6) -- cycle;
\draw (-1.3,1.4) -- (1.3,1.4) -- (1.3,0.6) -- (-1.3,0.6) -- cycle;
\draw (-0.3,0.4) -- (0.3,0.4) -- (0.3,-0.4) -- (-0.3,-0.4) -- cycle;
\end{tikzpicture}
} \qquad
\subfigure[]{
\begin{tikzpicture}[scale=0.7]
\tikzstyle{every node}=[draw,scale=0.6,fill=blue!20];
\path (0,3) node[ellipse] (c1) {$7$ $4$};
\path (0,2) node[ellipse] (c2) {$8$ $5$};
\path (1,2) node[ellipse] (c3) {$1$};
\path (0,1) node[ellipse] (c4) {$9$ $6$ $2$};
\path (0,0) node[ellipse] (c5) {$3$};
\tikzstyle{every node}=[scale=0.6];
\draw (c1) -- (c2) -- (c4) -- (c5);
\draw (c1) -- (c3);
\end{tikzpicture}
}
\caption{{{(a) Original estimates of the clusters in the partial grid using $V_1 = \{7,4\}$ as the root cluster. (b) Splitting of clusters. (c) Final block-tree. }}}
\label{fig:partialgrid-alternative}
\end{center}
\end{figure}
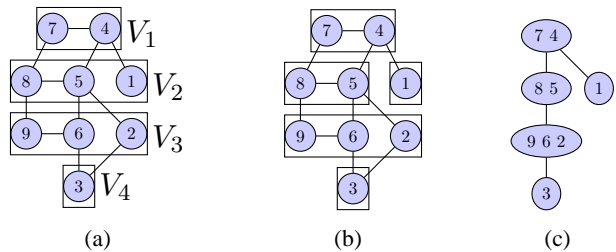

As an example illustrating how the choice of $V_1$ alters the block-tree, consider finding a block-tree for the partial grid in Fig.~\ref{fig:partialgrid-bt}(a).  In Fig.~\ref{fig:partialgrid-bt}(c), we constructed a block-tree using $V_1 = \{7\}$ as the root cluster.  The maximum sum of adjacent cluster sizes in Fig.~\ref{fig:partialgrid-bt}(c) is four.  Instead of choosing $V_1 = \{7\}$, let $V_1 = \{7,4\}$.  The initial estimate of the clusters are shown in Fig.~\ref{fig:partialgrid-alternative}(a).  The final block-tree is shown in Fig.~\ref{fig:partialgrid-alternative}(c).  Since the clusters $\{9,6,2\}$ and $\{8,5\}$ are adjacent, the maximum sum of adjacent clusters is five.

\subsection{Greedy Algorithms for Finding Optimal Block-Trees}
\label{sec:greedybt}

In the previous Section, we saw that finding optimal block-trees is computationally intractable.  In this Section, we propose three greedy algorithms for finding optimal block-trees that have varying degrees of computational complexity.

\noindent
\textbf{Minimal degree node - $\MinDegree$:} In this approach, which we call $\MinDegree$, we find the node with minimal degree and use that node as the root cluster.  The intuition behind this is that the minimal degree node may lead to the smallest number of nodes being added in the clusters.  The complexity of this approach is $O(n)$, where $n$ is the number of nodes in the graph.

The next two algorithms are based on the relationship between junction trees and block-trees outlined in Section~\ref{sec:optimalbt}.  Recall that for every block-tree, we can find a junction tree. This means that an optimal junction tree (a junction tree with minimal width) may be used to find an approximate optimal block-tree.  Further, finding optimal junction trees corresponds to finding optimal elimination orders in graphs \cite{JensenJenson1994}.  Thus, we can make use of greedy algorithms for finding optimal elimination orders to find optimal block-trees.

\noindent
\textbf{Using an elimination order - $\GreedyDegree$:}  One of the simplest algorithms for finding an approximate optimal elimination order is known as $\GreedyDegree$ \cite{Markowitz1957,BerryElimination2003}, where the elimination order corresponds to the sorted list of nodes in increasing degree.  The complexity of $\GreedyDegree$ is $O(n \log n)$ since we just need to sort the nodes.  Using the elimination order, we triangulate\footnote{A graph is triangulated if each cycle of length four or more has an edge connecting non adjacent nodes.} the graph to find the cliques.  
These cliques correspond to the set of clusters in the junction tree representation.  We search over a constant number of cliques to find an optimal root cluster.

\noindent
\textbf{Using an elimination order - $\GreedyFillin$:} Another popular greedy algorithm is to find an optimal elimination order such that at each step in the triangulation algorithm, we choose a node that adds a minimal number of extra edges in the graph.  This is known as $\GreedyFillin$ \cite{Kjaerulff1990} and has polynomial complexity.  Thus, $\GreedyFillin$ is in general slower than $\GreedyDegree$, but does lead to slightly better elimination orders on average.  To find the block-tree, we again search over a constant number of cliques over the triangulated graph.

\begin{figure}
\begin{center}
\includegraphics[scale=0.4]{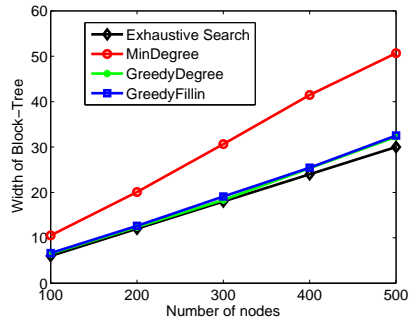}
\end{center}
\caption{Plot showing the performance of three different greedy heuristics for finding optimal block-trees.}
\label{fig:blocktreeapprox}
\end{figure}

We now evaluate the three different greedy algorithms, $\MinDegree$, $\GreedyDegree$, and $\GreedyFillin$, for finding optimal block-trees in Fig.~\ref{fig:blocktreeapprox}.  To do this, we create clusters of size $k$ such that the total number of nodes is $n$ (one cluster may have less than $k$ nodes).  We then form a tree over the clusters and associate a clique between two clusters connected to each other.  We then remove a certain fraction of edges over the graph (not the block-tree), but make sure that the graph is still connected.  By construction, the width of the graph constructed is at most $2k$.  Fig.~\ref{fig:blocktreeapprox} shows the performance of $\MinDegree$, $\GreedyDegree$, and $\GreedyFillin$ over graphs with different number of nodes and different values of $k$.  We clearly see that both $\GreedyDegree$ and $\GreedyFillin$ compute widths that are close to optimal.  The main idea in this Section is that we can use various known algorithms for finding optimal junction trees to find optimal block-trees.

\subsection{Exact Inference Using Block-Trees}

In the literature, exact inference over graphical models using non-overlapping clusters is referred to as the Pearl's clustering algorithm \cite{Pearl1988}. In \cite{Cooper1984} and \cite{PengReggia1986}, the authors use non-overlapping clustering for some particular directed graphical models for an application in medical diagnostics.  For lattices, \cite{Woods1977,MouraBalram1992,LevyAdamsWillsky1990} derive inference algorithms by scanning the lattice horizontally (or vertically).  Our block-tree construction algorithm provides a principled way of finding non-overlapping clusters over arbitrary graphs.  

Inference over graphical models defined on block-trees can be done by extending the belief propagation (BP) algorithm~\cite{Pearl1988}.  
The computational complexity of BP will depend on $\max_{(i,j) \in {\cal E}} (|V_i| + |V_j|)$.  On the other hand, the computational complexity of exact inference using other frameworks that use overlapping clusters depends on the width of the graph \cite{LauritzenSpiegelhalter1988,ShaferShenoy1990,ZhangPoole1994,Dechter1999}, which is in general less than or equal to $\max_{(i,j) \in {\cal E}} (|V_i| + |V_j|)$.  The main advantage of using block-trees for exact inference is that the complexity of constructing block-trees is $O(|E|)$, whereas the complexity of constructing tree-decompositions for inference using frameworks that use overlapping clusters is worse than $O(|E|)$ \cite{JensenJenson1994,Dechter1999,KaskDechter2001}.  Thus, block-trees are suitable for exact inference over time-varying graphical models \cite{kolar-song-xing-aoas09,ZhouWassLaff2010} such that the clusters in the block-tree are small.

\section{Block-Graph: Splitting Clusters in a Block-Tree}
\label{sec:blockgraph}

In this Section, we outline a greedy algorithm for splitting large clusters in a block-tree to form a block-graph.  This is an important step in our proposed framework (see Fig.~\ref{fig:generalizedAlg}) for generalizing inference algorithms since we apply the inference algorithm to the block-graph as opposed to the original graph.  Note that we can use the block-tree itself for inference; however, for many graphs this is computationally intractable since the complexity of inference for discrete graphical models using a block-tree is exponential in $\max_{(i,j) \in {\cal E}} (|V_i| + |V_j|)$.  Thus, when the size of one cluster in the block-tree is large, exact inference using block-trees will be computationally intractable\footnote{Our algorithm for finding optimal block-trees uses the junction tree construction algorithm, so even if $\max_{(i,j) \in {\cal E}} (|V_i| + |V_j|)$ is large and the treewidth of the graph is small, we can detect this and use junction trees for inference.}.

We modify Algorithm~\ref{alg:blocktree} for constructing block-trees to construct block-graphs such that all clusters have cardinality at most $m$.

\begin{enumerate}[Step 1.]
\item Using an initial cluster of nodes $V_1$, find clusters $V_1,V_2,\ldots,V_r$ using breadth-first search (BFS) such that $V_2 = {\cal N}(V_1), V_3 = {\cal N}(V_2) \backslash \{V_1 \cup V_2\},\ldots,V_r = {\cal N}(V_r) \backslash \{V_{r-2} \cup V_{r-1}\}$.  While doing the BFS, write $V_k$ as the set of all connected components in the subgraph $G(V_k)$.  Thus, $V_k$ is a set of clusters.

\item For $V_r$, if there exists any cluster that has cardinality greater than $m$, partition those components.  Let $V_r = \{V_r^1,V_r^2,\ldots,V_r^{m_r}\}$ be the final set of clusters.

\item Perform the next steps for each $k = r-1,r-2,\ldots, 1$, starting at $k = r-1$. Let $\widetilde{V_k}$ be the set of all clusters $V_k$ that have cardinality greater than $m$.  Partition all clusters in $\widetilde{V_k}$ into appropriate size clusters of size at most $m$.

\item Merge the clusters in the set $V_k \backslash \widetilde{V}_k$.  The idea used in merging clusters is that if two clusters are connected to the same cluster in $V_{k+1}$, by merging these two clusters, we reduce one edge in the final block-graph.  Further, if two clusters in $V_k$ are not connected to the same cluster in $V_{k+1}$, we do not merge these two clusters, since the number of edges in the final block-graph will remain the same.  The final clusters constructed using the above rules is denoted as $V_k = \{V_k^1,\ldots,V_k^{m_k}\}$.

\item The block-graph is given by the clusters 
$\displaystyle{{\cal V} =\{V_k^1,V_k^2,\ldots,V_k^{m_k}} \}_{k=1,\ldots,r}$ and the set of edges ${\cal E}$ between clusters.
\end{enumerate}

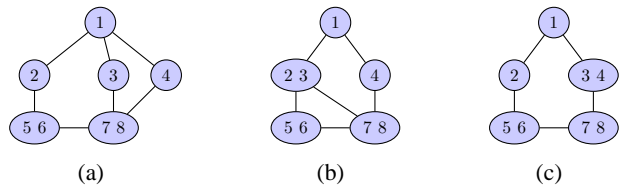
\begin{figure}
\begin{center}
\subfigure[]{
\begin{tikzpicture}[scale=0.7]
\tikzstyle{every node}=[draw,shape=circle,scale=0.6, fill=blue!20];
\path (0.75,2) node (t1) {$1$};
\path (-0.5,1) node (x1) {$2$};
\path (1,1) node (x2) {$3$};
\path (2,1) node (x3) {$4$};
\tikzstyle{every node}=[draw,shape=ellipse,scale=0.6, fill=blue!20];
\path (-0.5,0) node (c1) {$5$ $6$};
\path (1,0) node (c2) {$7$ $8$};
\draw (x1) -- (c1) -- (c2) -- (x2);
\draw (c2) -- (x3);
\draw (x1) -- (t1);
\draw (x2) -- (t1);
\draw (x3) -- (t1);
\end{tikzpicture}
} \qquad
\subfigure[]{
\begin{tikzpicture}[scale=0.7]
\tikzstyle{every node}=[draw,shape=circle,scale=0.6, fill=blue!20];
\path (0.25,2) node (t1) {$1$};
\path (1,1) node (x3) {$4$};
\tikzstyle{every node}=[draw,shape=ellipse,scale=0.6, fill=blue!20];
\path (-0.5,1) node (x1) {$2$ $3$};
\path (-0.5,0) node (c1) {$5$ $6$};
\path (1,0) node (c2) {$7$ $8$};
\draw (x1) -- (c1) -- (c2) -- (x3);
\draw (x1) -- (c2);
\draw (x1) -- (t1) -- (x3);
\end{tikzpicture}
} \qquad
\subfigure[]{
\begin{tikzpicture}[scale=0.7]
\tikzstyle{every node}=[draw,shape=circle,scale=0.6, fill=blue!20];
\path (0.25,2) node (t1) {$1$};
\path (-0.5,1) node (x1) {$2$};
\tikzstyle{every node}=[draw,shape=ellipse,scale=0.6, fill=blue!20];
\path (1,1) node (x2) {$3$ $4$};
\path (-0.5,0) node (c1) {$5$ $6$};
\path (1,0) node (c2) {$7$ $8$};
\draw (x1) -- (c1) -- (c2) -- (x2);
\draw (x1) -- (t1) -- (x2);
\end{tikzpicture}
}
\caption{{{Explaining Step 4 in the block-graph construction algorithm.  Given the block-graph in (a), if we merge nodes $2$ and $3$, we get the block-graph in (b).  If we merge nodes $3$ and $4$, we get the block-graph in (c).  The block-graph in (c) has just one loop.}}}
\label{fig:step6}
\end{center}
\end{figure}

The key step in the above algorithm is Step 4, where we cluster nodes appropriately.  Fig.~\ref{fig:step6} explains the intuition behind merging clusters with an example.  Suppose, we use the block-graph construction algorithm up to Step 3 and now we want to merge clusters in $V_2 = \{2,3,4\}$.  If we ignore Step 4 and merge clusters randomly, we might get the block-graph in Fig.~\ref{fig:step6}(b) on merging nodes $2$ and $3$.  If we use Step 4, since nodes $3$ and $4$ are connected to the same node, we merge these to get the block-graph in Fig.~\ref{fig:step6}(c).  The graph in Fig.~\ref{fig:step6}(c) has a single cycle with five edges, whereas the graph in Fig.~\ref{fig:step6}(b) has two cycles of size four and three.  It has been observed that inference over graphs with longer cycles is more accurate than inference over graphs with shorter cycles \cite{tfabre-isit00}.  Thus, our proposed algorithm leads to block-graphs that are favorable for inference.

\section{Inference Using Block-Graphs}
\label{sec:approximateinference}

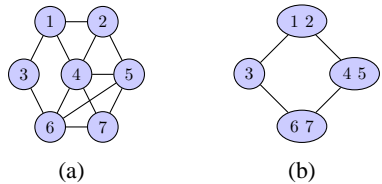
\begin{figure}
\begin{center}
\subfigure[]{
\begin{tikzpicture}[scale=0.7]
\tikzstyle{every node}=[draw,shape=circle,scale=0.6, fill=blue!20];
\path (-0.5,2) node (x1) {$1$};
\path (+0.5,2) node (x2) {$2$};
\path (-1,1) node (x3) {$3$};
\path (0,1) node (x4) {$4$};
\path (1,1) node (x5) {$5$};
\path (-0.5,0) node (x6) {$6$};
\path (0.5,0) node (x7) {$7$};
\draw (x1) -- (x2) -- (x5) -- (x4) -- (x1);
\draw (x2) -- (x4) -- (x6) -- (x7) -- (x4);
\draw (x7) -- (x5) -- (x6) -- (x3) -- (x1);
\end{tikzpicture}
} \qquad
\subfigure[]{
\begin{tikzpicture}[scale=0.7]
\tikzstyle{every node}=[draw,shape=circle,scale=0.6, fill=blue!20];
\path (-1,1) node (t1) {$3$};
\tikzstyle{every node}=[draw,shape=ellipse,scale=0.6, fill=blue!20];
\path (0,2) node (t0) {$1$ $2$};
\path (1,1) node (t2) {$4$ $5$};
\path (0,0) node (c2) {$6$ $7$};
\draw (t1) -- (t0) -- (t2) -- (c2) -- (t1);
\end{tikzpicture}
}
\caption{(a). Original graph. (b) Block-graph representation of the graph in (a).}
\label{fig:exampleinbg}
\end{center}
\end{figure}

Define a graphical model on a graph $G = (V,E)$ using a collection of $p$ random variables $\xb = (x_1,\ldots,x_p)$, where each $x_k$ takes values in $\Omega^d$, where $d \ge 1$.  Let ${\cal G} = ({\cal V},{\cal E})$ be a block-graph representation of the graph $G = (V,E)$.  To derive inference algorithms over the block-graph, we need to define appropriate potentials (or factors) associated with each clique in the block-graph.  This can be done by mapping the potentials from the original graph to the block-graph.  As an example, let $G$ be the graph in Fig.~\ref{fig:exampleinbg}(a) and let the probability distribution over $G$ be given by
\begin{equation}
 p(\xb) = \frac{1}{Z} \psi_{1,2}(x_1,x_2) \psi_{1,4}(x_1,x_4) \psi_{1,3}(x_1,x_3) \psi_{2,4,5}(x_2,x_4,x_5) \psi_{4,5,6,7}(x_4,x_5,x_6,x_7) \psi_{3,6}(x_3,x_6) 
\,.
\label{eq:pot}
\end{equation}
Let the clusters in the block-graph representation of $G$ in Fig.~\ref{fig:exampleinbg}(b) be $V_1 = \{1,2\}$, $V_2 = \{4,5\}$, $V_3 = \{3\}$, and $V_4 = \{6,7\}$.  The probability distribution in (\ref{eq:pot}) can be written in terms of the block-graph as follows:
\begin{equation}
 p(\xb) = \frac{1}{Z} \Psi_{1,2}(x_{V_1},x_{V_2}) \Psi_{1,3}(x_{V_1},x_{V_3}) \Psi_{2,4}(x_{V_2},x_{V_4}) \Psi_{2,4}(x_{V_2},x_{V_4}) 
\psi_{3,4}(x_{V_3},x_{V_4}) \,,
\label{eq:potbg}
\end{equation}
where
\begin{align}
\Psi_{1,2}(x_{V_1},x_{V_2}) &= \psi_{1,2}(x_1,x_2) \psi_{1,4}(x_1,x_4) \psi_{2,4,5}(x_2,x_4,x_5) \label{eq:pot1}\\
\Psi_{1,3}(x_{V_1},x_{V_3}) &= \psi_{1,3}(x_1,x_3) \\
\Psi_{2,4}(x_{V_2},x_{V_4}) &= \psi_{4,5,6,7}(x_4,x_5,x_6,x_7) \\
\Psi_{2,4}(x_{V_3},x_{V_4}) &= \psi_{3,6}(x_3,x_6) \label{eq:pot4} \,.
\end{align}

Let $\Alg$ be an algorithm for inference over graphical models.  Inference over the graph $G$ can be performed using $\Alg$ with inputs being the potentials in~(\ref{eq:pot}).  Inference over the block-graph can be performed using $\Alg$ with input being the potentials in (\ref{eq:pot1})-(\ref{eq:pot4}).  To get the marginal distributions from the block-graph, we need to further marginalize the joint probability distribution over each cluster.

\begin{remark}
 Both the representations (\ref{eq:pot}) and (\ref{eq:potbg}) are equivalent, so we are not making any approximations when parameterizing the graphical model using block-graphs.
\end{remark}

\begin{remark}
There is a trade-off in choosing the size of the clusters in the block-graph.  Generally, as observed in our numerical simulations, larger clusters lead to better estimates at the cost of more computations.
\end{remark}

\begin{remark}
 We presented the block-graph framework using undirected graphical models.  The results can be easily generalized to settings where the probability distribution is represented as a factor graph \cite{FactorGraphFrey2001}.
\end{remark}

\section{Numerical Simulations}
\label{sec:simulations}

In this Section, we provide numerical simulations to show how our proposed block-graph framework for generalizing inference algorithms can be used to improve the performance of current approximate inference algorithms that have been proposed in the literature.  Throughout this Section, we assume $x_s \in \{-1,+1\}$ and the probability distribution over $\xb$ factorizes as
\begin{equation}
 p(\xb) = \frac{1}{Z} \prod_{i = 1}^{p} \phi_i(x_i) \prod_{(i,j) \in E} \psi_{ij}(x_i,x_j) \,. \label{eq:isingpot}
\end{equation} 
The node potentials are given by $\phi_i(x_i) = \exp(-a_i x_i)$, where $a_i \sim {\cal N}(0,0.1)$ and the edge potentials are given by
\begin{align}
\text{Repulsive (REP): } \psi_{ij}(x_i,x_j) &= \exp(-|b_{ij}| x_i x_j ) \\
\text{Attractive (ATT): } \psi_{ij}(x_i,x_j) &= \exp(|b_{ij}| x_i x_j ) \\
\text{Mixed (MIX): } \psi_{ij}(x_i,x_j) &= \exp( - b_{ij} x_i x_j ) \,,
\end{align}
where $b_{ij} \sim {\cal N}(0,\sigma)$ and $\sigma$ is the interaction strength. For distributions with attractive (repulsive) potentials, neighboring random variables are more likely to take the same (opposite) value.  For distributions with mixed potentials, some neighbors are attractive, whereas some are repulsive.  We study several approximate inference algorithms that have been proposed in the literature: Belief Propagation (BP) \cite{Pearl1988}, Iterative Join-Graph Propagation (IJGP-$i$) \cite{MateescuKGDIJGP2010}, Generalized Belief Propagation (GBP-$i$) \cite{HAK2003,YedidaTIT2005}, Conditioned Belief Propagation (CBP-$l$) \cite{Eaton09}, Loop Corrected Belief Propagation (LC) \cite{MooijLC2007}, Tree-Structured Expectation Propagation (TreeEP) \cite{Minka03tree}.  In IJGP-$i$ and GBP-$i$, the integer $i$ refers to the maximum size of the clusters, where the clusters in these algorithms are overlapping.  The clusters in GBP-$i$ are selected by finding cycles of length $i$ in the graph.  In CBP-$l$, $l$ is an integer that refers to the number of clamped variables when performing inference: larger $l$ in general leads to more accurate marginal estimates.  We use the libDAI software package \cite{Mooij_libDAI_10} for all the inference algorithms except for IJGP, where we use the software provided by the authors at \cite{ijgpsoftware}.  For an inference algorithm Alg, we refer to the generalized inference algorithm as B$m$-Alg, where the $m$ is an integer denoting the maximum size of the cluster in the block-graph.

We consider two types of graphs: (i) grid graphs and (ii) random regular graphs, where each node in the graph has the same degree and the edges are chosen randomly.  Both these graphs have been used extensively in the literature for evaluating inference algorithms \cite{MooijLC2007,Eaton09}.  We compare inference algorithms using the mean absolute error:
\begin{equation}
 \text{Error} = \frac{1}{p} \sum_{s=1}^p \sum_{x_s \in \{-1,+1\}} |\widehat{p}_s(x_s) - p_s(x_s)| \,,
\end{equation}
where $\widehat{p}_s$ is the marginal estimate computed by an approximate inference algorithm and $p_s(x_s)$ is the true marginal distribution.  To evaluate the computational complexity, we measure the time taken in running the inference algorithms on a 2.66GHz Intel(R) Xeon(R) X5355 processor with 32 GB memory.
Since all the approximate inference algorithms we considered are iterative algorithms, we set the maximum number of iterations to be 1000 and stopped the inference algorithm when the mean absolute difference between the new and old marginal estimates is less than $10^{-9}$.  All the code and graphical models used in the numerical simulations can be downloaded from \url{http://www.ima.umn.edu/~dvats/GeneralizedInference.html}.

\subsection{Evaluating the Block-Graph Construction Algorithm}
\begin{figure}
\centering
 \includegraphics[scale=0.6]{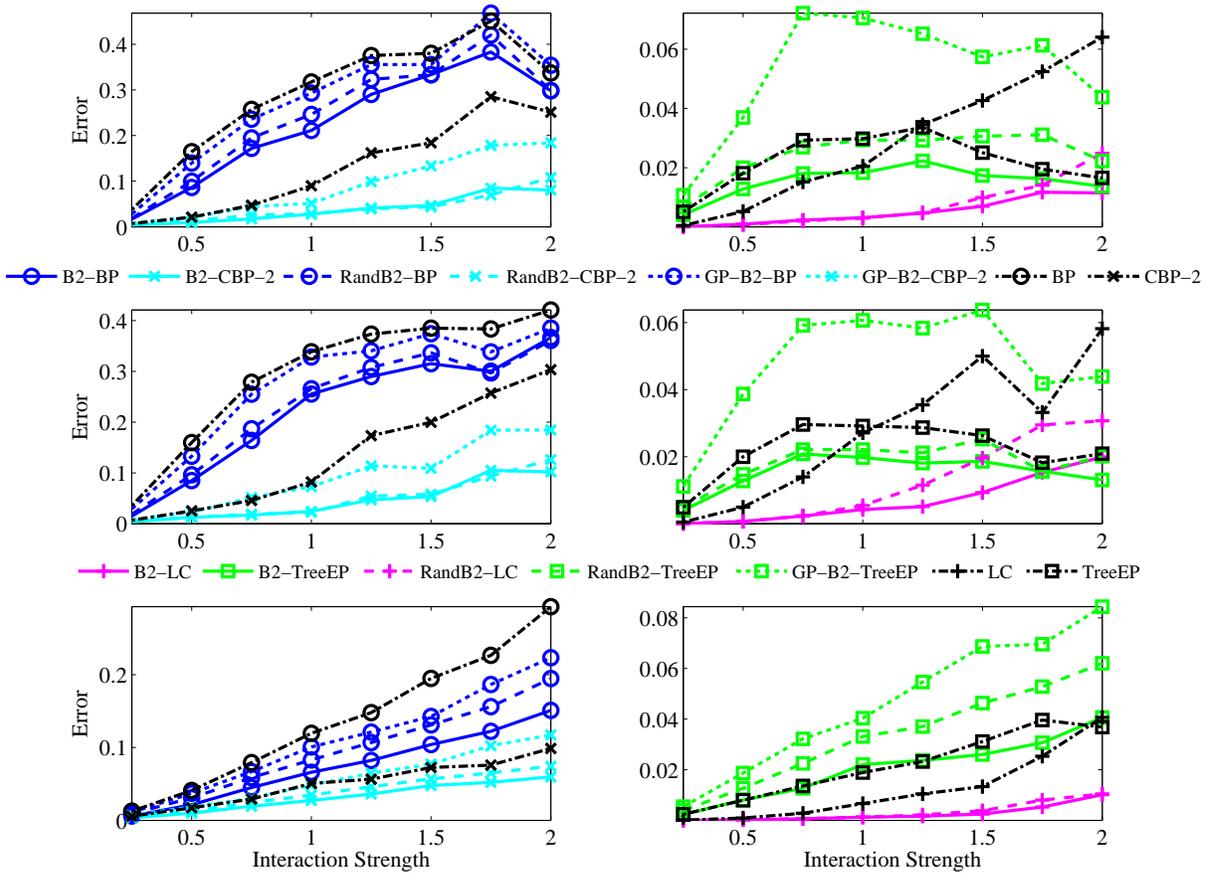}
\caption{Evaluating the block-graph construction algorithm in Section~\ref{sec:blockgraph} on a $5\times 5$ grid graph.  The solid lines, denoted by B$m$-Alg for an inference algorithm Alg, correspond to using our proposed block-graph construction algorithm.  The dashed lines correspond to an inference algorithm that randomly splits larger clusters in a block-tree.  The dotted lines correspond to an inference algorithm that uses graph partitioning to find clusters.  The plots in the top, middle, and bottom row correspond to repulsive, attractive, and mixed potentials, respectively.}
\label{fig:grid7}
\end{figure}

We first evaluate our proposed algorithm for constructing block-graphs (see Section~\ref{sec:blockgraph}) where we split large clusters in a block-tree.
Fig.~\ref{fig:grid7} shows the results of applying our block-graph framework to generalize BP, CBP, LC, and TreeEP on a $5\times 5$ grid graph.  We compare our algorithm to an algorithm that randomly splits the clusters in a block-tree and an algorithm proposed in \cite{XingGMFPartition2004} that uses graph partitioning to find non-overlapping clusters.  In Fig.~\ref{fig:grid7}, the solid lines correspond to our algorithm (see legend B2-BP), the dashed lines correspond to random splitting of clusters (see legend RandB2-BP), and the dotted lines correspond to graph partitioning (see legend GP-B2-BP).  The results reported are averages over $100$ trials.

\begin{remark}
It is clear that the graph partitioning approach performs the worst amongst the three different algorithms (the dotted line is above the solid and dashed line).  For TreeEP, we observe that the graph partitioning approach performs worse than the original algorithm that does not use block-graphs.  This suggests that the graph partitioning algorithm in \cite{XingGMFPartition2004} is not suitable for the inference algorithms considered in Fig.~\ref{fig:grid7}.  We did not apply the graph partitioning algorithm to LC since the corresponding inference algorithm was very slow.
\end{remark}

\begin{remark}
In most cases, our proposed algorithm for constructing block-graphs performs better than using an algorithm that randomly splits clusters (the solid line is below the dashed line).  Interestingly, for TreeEP, the random algorithm performs worse than the original algorithm.  We also observed that both the random algorithm and the graph partitioning algorithm took more time than our proposed algorithm.  This suggests that our proposed block-graph construction algorithm leads to block-graphs that are favorable for inference.
\end{remark}

\subsection{Grid Graphs}
Tables~\ref{tab:grid10rep}, \ref{tab:grid10att}, \ref{tab:grid10mix}, \ref{tab:grid15rep}, and \ref{tab:grid20rep} show results of applying the block-graph framework for inference over graphical models defined on grid graphs.  

\begin{figure}
\centering
 \includegraphics[scale=0.6]{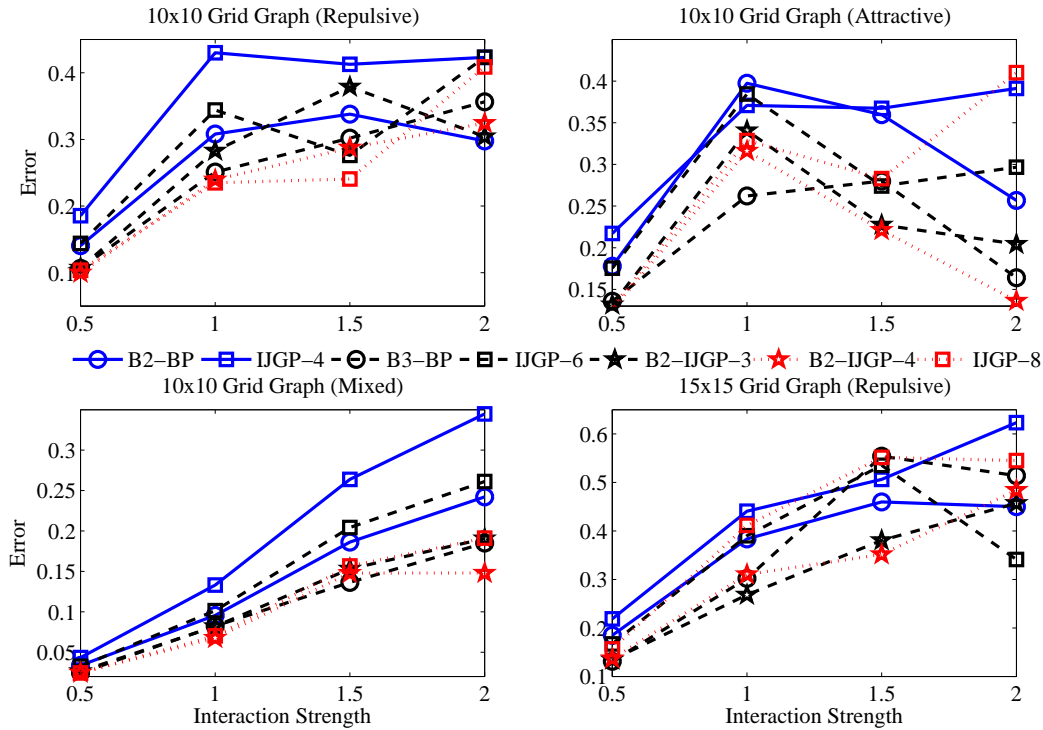}
\caption{Comparing the performance of BP, IJGP, and GBP.  Algorithms with the same message passing complexity are plotted in the same color.}
\label{fig:bpijgp}
\end{figure}

\begin{remark}
In general, we observe that for all cases considered, applying the block-graph framework leads to better marginal estimates.  This is shown in the Tables, where for each inference algorithm, we highlight the algorithm leading to the smallest mean error in bold.  For example, when using block-graphs of size two for BP, the error decreases by as much as $25\%$ (see BP vs. B$2$-BP), whereas when using block-graphs of size three for BP, the error decreases by as much as $50\%$ (see BP vs. B$3$-BP).
\end{remark}

\begin{remark}
It is interesting to compare BP, IJGP, and GBP, where both IJGP and GBP are based on finding overlapping clusters and IJGP first constructs a junction tree and then splits clusters in the junction tree to find overlapping clusters.  Note that for the class of graphical models considered in (\ref{eq:isingpot}), B$m$-BP belongs to the class of GBP-$2m$ algorithms since we can map the block-graph into an equivalent graph with overlapping clusters as done so when converting a block-tree into a junction tree (see Fig.~\ref{fig:btTojt}).  Further, IJBP-$2m$ is also a GBP-$2m$ algorithm \cite{MateescuKGDIJGP2010}.  Thus, we want to compare B$m$-BP, IJGP-$2m$, and BP-$2m$.  It is clear that GBP-$2m$ leads to superior marginal estimates, however, this comes at the cost of significantly more computations.  Fig.~\ref{fig:bpijgp} compares B$m$-BP to IJGP-$2m$.  We observe that for many cases, B$m$-BP leads to better marginal estimates than IJGP-$2m$.
We note that comparing B$m$-BP to IJGP-$2m$ may not be appropriate since the stopping criteria for the IJGP may be different than that of the BP algorithm\footnote{For IJGP, we used the software available at \cite{ijgpsoftware}.  We could specify the maximum number of iterations, but not the stopping criteria.}.
\end{remark}

\begin{remark}
 We can apply the block-graph framework to generalize GBP based algorithms.  Our results show that this leads to better marginal estimates, see GBP-4 vs. B2-GBP-4, IJGP-3 vs. B2-IJGP-3, and IJGP-4 vs. B2-IJGP-4.  More specifically, looking at Table~\ref{tab:grid20rep}, we notice that the performance of using block-graphs of size two on GBP results in the error being reduced to nearly $15\%$ of the original error.  
\end{remark}

\begin{remark}
In Fig.~\ref{fig:bpijgp}, we see that for many cases the performance of B2-IJGP-3 (B2-IJGP-4) is better than IJGP-6 (IJGP-8).  This suggests that the set of overlapping clusters chosen using the block-graph framework may be better than the clusters chosen using the IJGP framework.
\end{remark}

\begin{remark}
Overall, we observe that block-graph versions of TreeEP lead to the best estimates with reasonable computational time.  For example, in Table~\ref{tab:grid15rep}, with $\sigma = 1$, B2-TreeEP results in a mean error of $0.1583$ running in an average of $0.455$ seconds.  In comparison, GBP-$4$  takes an average of about $95$ seconds and the mean error is $0.0884$.  When compared to other algorithms, B$3$-CBP-$2$ runs in $0.32$ seconds and results in a mean error of $0.2657$.  For mixed potentials in Table~\ref{tab:grid10mix}, we observe that the generalized versions of TreeEP do not lead to significant improvements in the marginal estimates although the performance of other algorithms does improve.  Reference \cite{WelMinTeh2005} proposes a generalization of TreeEP and gives guidelines for choosing clusters in the GBP algorithm.  As shown for IJGP and GBP, our framework can be used in conjunction with frameworks that use overlapping clusters.
\end{remark}

\begin{remark}
To our knowledge, there have been no algorithms for generalizing LC and CBP-$l$.  The computational complexity of LC is exponential in the maximum degree of the graph \cite{MooijLC2007}, so it is only feasible to apply LC to a limited number of graphs.  We only used LC for the $5\times5$ grid graph example in Fig.~\ref{fig:grid7}.  We observe that the CBP-$l$ algorithm improves the estimates of the BP algorithm.  Moreover, for regimes where the interaction strength is small, the performance of generalized versions of CBP is comparable to that of TreeEP.  For example, in Table~\ref{tab:grid15rep},  for $\sigma=0.5$, the best TreeEP algorithm has a mean error of $0.0694$ and the best CBP based algorithm has a mean error of $0.0864$.  As another example, in Table~\ref{tab:grid20rep}, for $\sigma = 0.5$, the best TreeEP algorithm has a mean error of $0.0624$ and the best CBP algorithm has a mean error of $0.0608$.
\end{remark}

\begin{figure}
\centering
\subfigure[$\sigma=0.5$]{
 \includegraphics[scale=0.5]{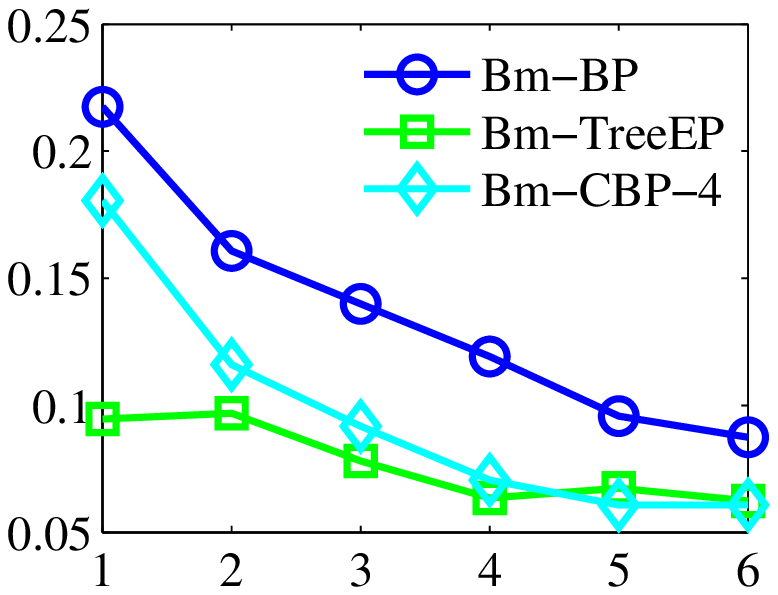}
}
\subfigure[$\sigma=1.0$]{
 \includegraphics[scale=0.5]{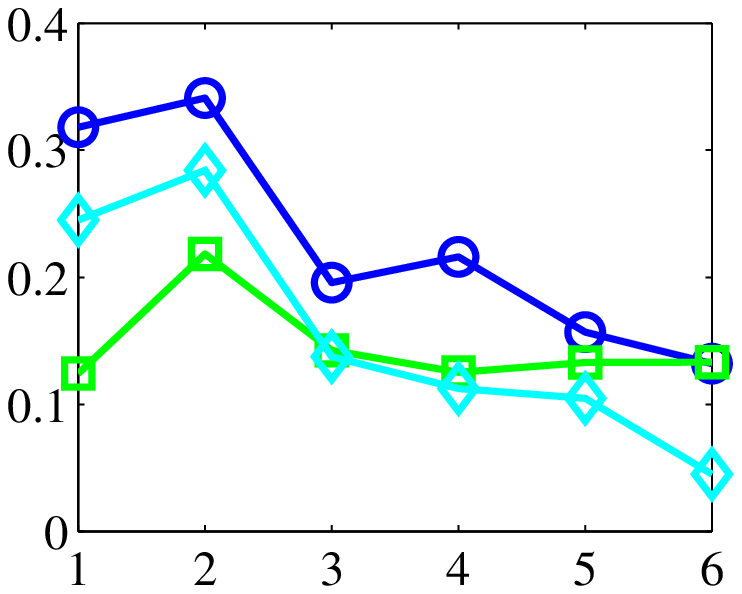}
}
\subfigure[$\sigma=1.5$]{
 \includegraphics[scale=0.5]{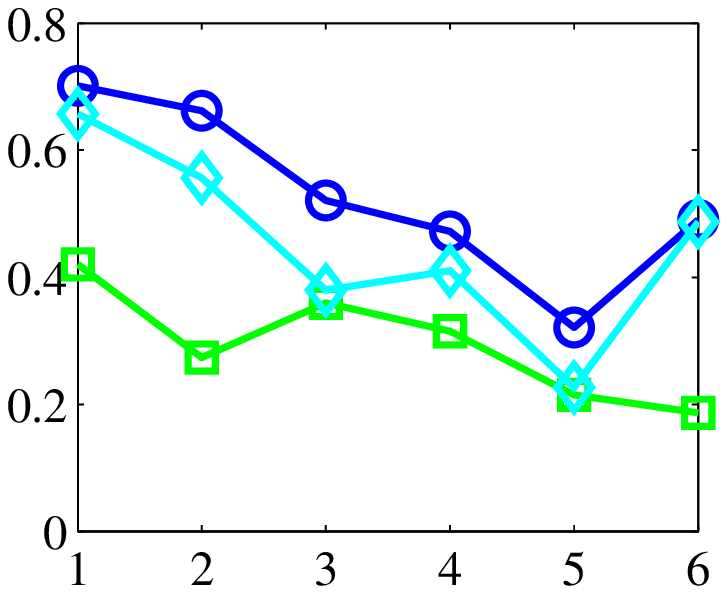}
}
\caption{Error as the size of the cluster increases in the $20 \times 20$ grid graph.  The horizontal axis denotes the cluster size in the block-graph and vertical axis denotes the mean error.}
\label{fig:blocksize}
\end{figure}

\begin{remark}
 Fig.~\ref{fig:blocksize} shows how the error scales as the size of the cluster in the block-graph increases for the $20\times 20$ grid graph.  It is clear that the error in general decreases as the cluster size increases; however, for some cases, the error does seem to increase especially when the interaction strength is large.
\end{remark}

\begin{table}
\caption{$10\times 10$ Grid Graph with Repulsive Potentials: $30$ Trials}
\label{tab:grid10rep}
\centering
{\myfontsize{
   \begin{tabular}{l r r c r r c r r c r r} \toprule
        \multirow{2}{*}{ Algorithm} & \multicolumn{2}{c}{$\sigma = 0.5$} && \multicolumn{2}{c}{$\sigma = 1$} && 
	\multicolumn{2}{c}{$\sigma = 1.5$} && \multicolumn{2}{c}{$\sigma = 2.0$}\\
	  \cmidrule{2-3} \cmidrule{5-6} \cmidrule{8-9} \cmidrule{11-12}
        & 
        \multicolumn{1}{c}{Error} & 
        \multicolumn{1}{c}{Time (s)} &&
        \multicolumn{1}{c}{Error} &
        \multicolumn{1}{c}{Time (s)} &&
        \multicolumn{1}{c}{Error} &
        \multicolumn{1}{c}{Time (s)} &&
        \multicolumn{1}{c}{Error} &
        \multicolumn{1}{c}{Time (s)}  \\
        \hline
BP &     {0.2122} &     0.0457 &&     {0.3714} &     0.0237 &&     {0.4773} &     0.0150 &&     {0.4220 }&     0.0120 \\
B2-BP &     0.1405 &     0.0213 &&     0.3080 &     0.0073 &&     0.3379 &     0.0057 &&     0.2978 &     0.0037 \\
B3-BP &     \textbf{0.1065} &     0.0193 &&     \textbf{0.2509} &     0.0133 &&     \textbf{0.3019} &     0.0033 &&     \textbf{0.3565} &     0.0020 \\
\hline
IJGP-3 &     0.1864 &        - &&     0.3784 &        - &&     0.4560 &        - &&     0.4254 &        - \\
B2-IJGP-3 &     0.1073 &        - &&     0.2827 &        - &&     0.3789 &        - &&     \textbf{0.3044} &        - \\
IJGP-6 &     0.1441 &        - &&     0.3442 &        - &&     0.2761 &        - &&     0.4232 &        - \\
IJGP-4 &     0.1856 &        - &&     0.4300 &        - &&     0.4128 &        - &&     0.4230 &        - \\
B2-IJGP-4 &     \textbf{0.0997} &        - &&     0.2394 &        - &&     0.2873 &        - &&     0.3245 &        - \\
IJGP-8 &     0.1038 &        - &&     \textbf{0.2349} &        - &&     \textbf{0.2407 }&        - &&     0.4088 &        - \\
\hline
CBP-2 &     0.1345 &     0.2507 &&     0.2757 &     0.1710 &&     0.4405 &     0.1240 &&     0.3801 &     0.1137 \\
B2-CBP-2 &     0.0740 &     0.1687 &&     0.2109 &     0.1147 &&     0.3263 &     0.0870 &&     0.2871 &     0.0660 \\
B3-CBP-2 &     0.0490 &     0.1583 &&     0.1872 &     0.1213 &&     0.2701 &     0.0890 &&     0.3756 &     0.0767 \\
CBP-3 &     0.1056 &     0.5223 &&     0.2224 &     0.3537 &&     0.4176 &     0.2603 &&     0.3459 &     0.2313 \\
B2-CBP-3 &     0.0561 &     0.3470 &&     0.1726 &     0.2483 &&     0.2824 &     0.1863 &&     0.2752 &     0.1470 \\
CBP-4 &     0.0866 &     1.0303 &&     0.2094 &     0.6913 &&     0.3420 &     0.5337 &&     0.3195 &     0.4650 \\
B2-CBP-4 &     \textbf{0.0437} &     0.6810 &&     \textbf{0.1229} &     0.5067 &&     \textbf{0.2438} &     0.3800 &&     \textbf{0.2744} &     0.3123 \\
\hline
TreeEP &     0.0678 &     0.1993 &&     0.1499 &     0.2513 &&     0.1475 &     0.1680 &&     0.1067 &     0.1630 \\
B2-TreeEP &     0.0547 &     0.1110 &&     0.1273 &     0.1343 &&     0.0489 &     0.1400 &&     0.0551 &     0.1120 \\
B3-TreeEP &     \textbf{0.0542} &     0.1463 &&     \textbf{0.0878} &     0.1683 &&     \textbf{0.0485} &     0.1447 &&     \textbf{0.0414} &     0.1527 \\
\hline
GBP-4 &     0.0110 &    21.0497 &&     0.0439 &    28.8153 &&     0.0532 &    25.7290 &&     0.0379 &    25.5437 \\
B2-GBP-4 &     \textbf{0.0005} &    16.7593 &&     \textbf{0.0026} &    25.5223 &&     \textbf{0.0021} &    28.8153 &&     \textbf{0.0018} &    33.1343 \\
    \bottomrule
    \end{tabular}
}}
\end{table}

{{
\begin{table}
\centering
{\myfontsize{
\caption{$10\times 10$ Grid Graph with Attractive Potentials: $30$ Trials}
\label{tab:grid10att}
   \begin{tabular}{l r r c r r c r r c r r} \toprule
        \multirow{2}{*}{Algorithm} & \multicolumn{2}{c}{$\sigma = 0.5$} && \multicolumn{2}{c}{$\sigma = 1$} && 
	\multicolumn{2}{c}{$\sigma = 1.5$} && \multicolumn{2}{c}{$\sigma = 2.0$}\\
	  \cmidrule{2-3} \cmidrule{5-6} \cmidrule{8-9} \cmidrule{11-12}
        & 
        \multicolumn{1}{c}{Error} & 
        \multicolumn{1}{c}{Time (s)} &&
        \multicolumn{1}{c}{Error} &
        \multicolumn{1}{c}{Time (s)} &&
        \multicolumn{1}{c}{Error} &
        \multicolumn{1}{c}{Time (s)} &&
        \multicolumn{1}{c}{Error} &
        \multicolumn{1}{c}{Time (s)}  \\
        \hline
BP &     {0.2337} &     0.0600 &&     0.4482 &     0.0217 &&     {0.3857} &     0.0160 &&     {0.3537} &     0.0113 \\
B2-BP &     0.1778 &     0.0227 &&     0.3975 &     0.0080 &&     0.3597 &     0.0083 &&     0.2567 &     0.0053 \\
B3-BP &     \textbf{0.1358} &     0.0173 &&     \textbf{0.2622} &     0.0080 &&     \textbf{0.2799} &     0.0090 &&     \textbf{0.1640} &     0.0027 \\
\hline
IJGP-3 &     0.2125 &        - &&     0.3710 &        - &&     0.3088 &        - &&     0.3232 &        - \\
B2-IJGP-3 &     0.1316 &        - &&     0.3406 &        - &&     0.2275 &        - &&     0.2044 &        - \\
IJGP-6 &     0.1755 &        - &&     0.3846 &        - &&     0.2741 &        - &&     0.2967 &        - \\
IJGP-4 &     0.2171 &        - &&     0.3708 &        - &&     0.3674 &        - &&     0.3912 &        - \\
B2-IJGP-4 &     \textbf{0.1259} &        - &&     \textbf{0.3161} &        - &&     \textbf{0.2211} &        - &&     \textbf{0.1360} &        - \\
IJGP-8 &     0.1291 &        - &&     0.3287 &        - &&     0.2831 &        - &&     0.4102 &        - \\
\hline
CBP-2 &     0.1468 &     0.2543 &&     0.3710 &     0.1590 &&     0.3389 &     0.1337 &&     0.3252 &     0.1007 \\
B2-CBP-2 &     0.0687 &     0.1647 &&     0.2913 &     0.1127 &&     0.3177 &     0.0877 &&     0.2521 &     0.0707 \\
B3-CBP-2 &     0.0506 &     0.1607 &&     0.1979 &     0.1143 &&     0.2539 &     0.0943 &&     \textbf{0.1092} &     0.0763 \\
CBP-3 &     0.1028 &     0.5243 &&     0.3289 &     0.3300 &&     0.2648 &     0.2713 &&     0.2855 &     0.2180 \\
B2-CBP-3 &     0.0490 &     0.3423 &&     0.2522 &     0.2400 &&     0.3006 &     0.1893 &&     0.2416 &     0.1577 \\
CBP-4 &     0.0784 &     1.0373 &&     0.2595 &     0.6690 &&     \textbf{0.2182} &     0.5497 &&     0.2662 &     0.4593 \\
B2-CBP-4 &     \textbf{0.0382} &     0.6737 &&     \textbf{0.1839 }&     0.4723 &&     0.2847 &     0.3857 &&     0.1682 &     0.3233 \\
\hline
TreeEP &     0.0804 &     0.2300 &&     0.2153 &     0.2470 &&     0.1128 &     0.1720 &&     0.0803 &     0.1977 \\
B2-TreeEP &     0.0499 &     0.1327 &&     0.1390 &     0.1787 &&     0.1104 &     0.1133 &&     0.0552 &     0.1020 \\
B3-TreeEP &    \textbf{ 0.0427} &     0.1407 &&     \textbf{0.0989} &     0.2117 &&     \textbf{0.0686 }&     0.2090 &&    \textbf{ 0.0546 }&     0.1337 \\
\hline
GBP-4 &     0.0085 &    21.4370 &&     0.0500 &    30.0877 &&     0.0501 &    25.9627 &&     0.0340 &    24.5683 \\
B2-GBP-4 &     \textbf{0.0005} &    16.7007 &&     \textbf{0.0033 }&    26.8110 &&    \textbf{ 0.0022} &    28.9640 &&     \textbf{0.0012} &    32.2330 \\
    \bottomrule
    \end{tabular}
}}
\end{table}
}}

\begin{table}
\centering
{\myfontsize{
\caption{$10\times 10$ Grid Graph with Mixed Potentials: $30$ Trials}
\label{tab:grid10mix}
   \begin{tabular}{l r r c r r c r r c r r} \toprule
        \multirow{2}{*}{Algorithm} & \multicolumn{2}{c}{$\sigma = 0.5$} && \multicolumn{2}{c}{$\sigma = 1$} && 
	\multicolumn{2}{c}{$\sigma = 1.5$} && \multicolumn{2}{c}{$\sigma = 2.0$}\\
	  \cmidrule{2-3} \cmidrule{5-6} \cmidrule{8-9} \cmidrule{11-12}
        & 
        \multicolumn{1}{c}{Error} & 
        \multicolumn{1}{c}{Time (s)} &&
        \multicolumn{1}{c}{Error} &
        \multicolumn{1}{c}{Time (s)} &&
        \multicolumn{1}{c}{Error} &
        \multicolumn{1}{c}{Time (s)} &&
        \multicolumn{1}{c}{Error} &
        \multicolumn{1}{c}{Time (s)}  \\
        \hline
BP &     {0.0514} &     0.0237 &&     {0.1542} &     0.2863 &&     {0.3178} &     0.4313 &&     {0.3728} &     0.4750 \\
B2-BP &     0.0337 &     0.0057 &&     0.0955 &     0.0310 &&     0.1862 &     0.1653 &&     0.2422 &     0.2320 \\
B3-BP &     \textbf{0.0243} &     0.0057 &&     \textbf{0.0824} &     0.0363 &&     \textbf{0.1364} &     0.1217 &&     \textbf{0.1851} &     0.1883 \\
\hline
IJGP-3 &     0.0431 &        - &&     0.1362 &        - &&     0.2719 &        - &&     0.3588 &        - \\
B2-IJGP-3 &     0.0264 &        - &&     0.0820 &        - &&     0.1531 &        - &&     0.1905 &        - \\
IJGP-6 &     0.0325 &        - &&     0.1015 &        - &&     0.2042 &        - &&     0.2612 &        - \\
IJGP-4 &     0.0434 &        - &&     0.1330 &        - &&     0.2639 &        - &&     0.3449 &        - \\
B2-IJGP-4 &     0.0246 &        - &&     \textbf{0.0675} &        - &&     \textbf{0.1479} &        - &&     \textbf{0.1479 }&        - \\
IJGP-8 &     \textbf{0.0243} &        - &&     0.0703 &        - &&     0.1567 &        - &&     0.1911 &        - \\
\hline
CBP-2 &     0.0391 &     0.2240 &&     0.0988 &     0.3727 &&     0.1789 &     0.3817 &&     0.2184 &     0.3913 \\
B2-CBP-2 &     0.0240 &     0.1190 &&     0.0692 &     0.2107 &&     0.1107 &     0.2583 &&     0.1572 &     0.2653 \\
B3-CBP-2 &     \textbf{0.0164} &     0.1230 &&     \textbf{0.0505} &     0.1957 &&     0.1020 &     0.2440 &&     0.1212 &     0.2600 \\
CBP-3 &     0.0351 &     0.4950 &&     0.0879 &     0.8083 &&     0.1549 &     0.8327 &&     0.1984 &     0.8497 \\
B2-CBP-3 &     0.0208 &     0.2660 &&     0.0642 &     0.4753 &&     0.0988 &     0.5790 &&     0.1325 &     0.6020 \\
CBP-4 &     0.0314 &     1.0173 &&     0.0780 &     1.6667 &&     0.1484 &     1.7300 &&     0.1755 &     1.7780 \\
B2-CBP-4 &     0.0191 &     0.5620 &&     0.0541 &     0.9970 &&     \textbf{0.0986} &     1.1933 &&     \textbf{0.1004} &     1.2360 \\
\hline
TreeEP &     0.0124 &     0.1190 &&     0.0350 &     0.1650 &&     0.0610 &     0.2177 &&     0.0864 &     0.2743 \\
B2-TreeEP &     \textbf{0.0124} &     0.0837 &&     0.0386 &     0.1353 &&     \textbf{0.0573} &     0.1467 &&     \textbf{0.0825} &     0.2340 \\
B3-TreeEP &     0.0125 &     0.1057 &&     \textbf{0.0342} &     0.1490 &&     0.0681 &     0.1980 &&     0.0887 &     0.2413 \\
\hline
GBP-4 &     0.0009 &    10.8840 &&     0.0054 &    17.2403 &&     0.0091 &    22.2760 &&     0.0139 &    23.1837 \\
B2-GBP-4 &     \textbf{0.0000} &    10.8890 &&     \textbf{0.0002} &    15.1400 &&    \textbf{ 0.0008} &    17.8040 &&     \textbf{0.0013} &    18.6687 \\

    \bottomrule
    \end{tabular}
}}
\end{table}

\begin{table}
\centering
{\myfontsize{
\caption{$15\times 15$ Grid Graph with Repulsive Potentials: $20$ Trials}
\label{tab:grid15rep}
   \begin{tabular}{l r r c r r c r r c r r} \toprule
        \multirow{2}{*}{Algorithm} & \multicolumn{2}{c}{$\sigma = 0.5$} && \multicolumn{2}{c}{$\sigma = 1$} && 
	\multicolumn{2}{c}{$\sigma = 1.5$} && \multicolumn{2}{c}{$\sigma = 2.0$}\\
	  \cmidrule{2-3} \cmidrule{5-6} \cmidrule{8-9} \cmidrule{11-12}
        & 
        \multicolumn{1}{c}{Error} & 
        \multicolumn{1}{c}{Time (s)} &&
        \multicolumn{1}{c}{Error} &
        \multicolumn{1}{c}{Time (s)} &&
        \multicolumn{1}{c}{Error} &
        \multicolumn{1}{c}{Time (s)} &&
        \multicolumn{1}{c}{Error} &
        \multicolumn{1}{c}{Time (s)}  \\
        \hline
BP &     {0.2187 }&     0.1550 &&     {0.3977 }&     0.0870 &&     {0.5307} &     0.0760 &&     {0.5737 }&     0.0575 \\
B2-BP &     0.1848 &     0.0745 &&     0.3836 &     0.0590 &&     0.4598 &     0.0400 &&     0.4500 &     0.0385 \\
B3-BP &     \textbf{0.1314} &     0.0700 &&    \textbf{ 0.3020 }&     0.0455 &&     \textbf{0.5541 }&     0.0360 &&     \textbf{0.5140} &     0.0445 \\
\hline
IJGP-3 &     0.2052 &        - &&     0.4239 &        - &&     0.4846 &        - &&     0.5706 &        - \\
B2-IJGP-3 &     \textbf{0.1346} &        - &&     \textbf{0.2683 }&        - &&     0.3805 &        - &&     0.4571 &        - \\
IJGP-6 &     0.1671 &        - &&     0.3903 &        - &&     0.5354 &        - &&     \textbf{0.3411} &        - \\
IJGP-4 &     0.2187 &        - &&     0.4406 &        - &&     0.5065 &        - &&     0.6234 &        - \\
B2-IJGP-4 &     0.1367 &        - &&     0.3103 &        - &&     \textbf{0.3515 }&        - &&     0.4842 &        - \\
IJGP-8 &     0.1566 &        - &&     0.4115 &        - &&     0.5518 &        - &&     0.5451 &        - \\
\hline
CBP-2 &     0.1774 &     0.6895 &&     0.3609 &     0.4940 &&     0.4802 &     0.4440 &&     0.5789 &     0.3435 \\
B2-CBP-2 &     0.1335 &     0.4550 &&     0.3161 &     0.3390 &&     0.4749 &     0.2705 &&     0.4517 &     0.2425 \\
B3-CBP-2 &     0.0887 &     0.4520 &&     \textbf{0.2657} &     0.3200 &&     0.5948 &     0.3115 &&     0.4910 &     0.2515 \\
CBP-3 &     0.1578 &     1.4245 &&     0.3234 &     1.0085 &&     0.4503 &     0.8665 &&     0.5113 &     0.7035 \\
B2-CBP-3 &     0.0946 &     0.9670 &&     0.2964 &     0.6650 &&     0.3852 &     0.5590 &&     0.4319 &     0.5080 \\
CBP-4 &     0.1382 &     2.9145 &&     0.2904 &     1.9985 &&     0.4150 &     1.6810 &&     0.5364 &     1.3610 \\
B2-CBP-4 &     \textbf{0.0864} &     1.9710 &&     0.2713 &     1.3210 &&     \textbf{0.3778} &     1.1395 &&     \textbf{0.3966} &     1.0225 \\
\hline
TreeEP &     0.0954 &     0.8500 &&     0.2831 &     0.7995 &&     0.5025 &     0.5895 &&     0.7151 &     0.4345 \\
B2-TreeEP &     0.0888 &     0.5265 &&     \textbf{0.1583} &     0.4550 &&     0.3566 &     0.4135 &&     0.4293 &     0.3530 \\
B3-TreeEP &     \textbf{0.0694} &     0.7530 &&     0.1600 &     0.7835 &&     \textbf{0.2584} &     0.5290 &&     \textbf{0.1375} &     0.6025 \\
\hline
GBP-4 &     0.0151 &    88.3395 &&     0.0884 &    95.5370 &&     0.0918 &    86.9865 &&     0.0639 &    88.9645 \\
B2-GBP-4 &     \textbf{0.0014 }&    81.1430 &&     \textbf{0.0120} &   117.4050 &&     \textbf{0.0105} &   117.1510 &&    \textbf{ 0.0062 }&   120.3190 \\
    \bottomrule
    \end{tabular}
}}
\end{table}

\begin{table}
\centering
{\myfontsize{
\caption{$20\times 20$ Grid Graph with Repulsive Potentials: $10$ Trials}
\label{tab:grid20rep}
\begin{tabular}{l r r r r || l r r r r } \toprule
{Algorithm} & {$\sigma = 0.5$} & {$\sigma = 1$} & 
{$\sigma = 1.5$} & {$\sigma = 2.0$} & {Algorithm} & {$\sigma = 0.5$} & {$\sigma = 1$} &
{$\sigma = 1.5$} & {$\sigma = 2.0$}\\
\hline

BP &     0.2174 &     0.3182 &     0.7014 &     0.5926 & CBP-2 &     0.2028 &     0.3222 &     0.7034 &     0.5550 \\
B2-BP &     0.1608 &     0.3412 &     0.6623 &     0.5117 & B2-CBP-2 &     0.1387 &     0.3023 &     0.6497 &     0.4465 \\
B6-BP &     \textbf{0.0874} &     \textbf{0.1320} &     \textbf{0.4904} &    \textbf{ 0.3102} & B6-CBP-2 &     0.0638 &     0.0952 &     0.5348 &     0.2926 \\
\hline
TreeEP &     0.0946 &     0.1243 &     0.4201 &     0.5124 & CBP-3 &     0.1893 &     0.3042 &     0.6634 &     0.5569 \\
B2-TreeEP &     0.0969 &     0.2184 &     0.2736 &     0.5154 & B2-CBP-3 &     0.1248 &     0.2385 &     0.5538 &     0.4057 \\
B6-TreeEP &     \textbf{0.0624} &    \textbf{ 0.1330} &     \textbf{0.1868} &     \textbf{0.0574} & B6-CBP-3 &     0.0637 &     0.0545 &     0.4947 &     \textbf{0.2919} \\
\hline
GBP-4 &     0.0175 &     0.0285 &     0.1188 &     0.0259 & CBP-4 &     0.1806 &     0.2450 &     0.6571 &     0.5409 \\
B2-GBP-4 &     0.0015 &     0.0034 &     0.0264 &     0.0038 & B2-CBP-4 &     0.1160 &     0.2842 &     0.5563 &     0.3990 \\
B3-GBP-4 &     \textbf{0.0003} &     \textbf{0.0005 }&     \textbf{0.0021 }&     \textbf{0.0017 }& B6-CBP-4 &     \textbf{0.0608} &     \textbf{0.0451} &     \textbf{0.4871} &     0.3116 \\
\bottomrule
\end{tabular}
}}
\end{table}

\subsection{Random Regular Graphs}

Tables~\ref{tab:randgraph50} and \ref{tab:randgraph70} show results of applying the block-graph framework for inference over graphical models defined on random regular graphs with attractive potentials.  Table~\ref{tab:randgraph50} considers graphs with $50$ nodes and degree $3$ and Table~\ref{tab:randgraph70} considers graphs with $70$ nodes and degree $3$.  Just like the grid graph case, we observe that our generalized framework leads to better marginal estimates.  This is shown by highlighting the algorithm that leads to minimal mean error for each inference algorithm.  We observe that both CBP and TreeEP based algorithms perform the best, even when compared to GBP.

\begin{table}
\centering
{\myfontsize{
\caption{Random Regular Graph with $p = 50$ Nodes, Degree $3$, and Attractive potentials: $30$ Trials}
\label{tab:randgraph50}
   \begin{tabular}{l r r c r r c r r c r r} \toprule
        \multirow{2}{*}{Algorithm} & \multicolumn{2}{c}{$\sigma = 0.5$} && \multicolumn{2}{c}{$\sigma = 1$} && 
	\multicolumn{2}{c}{$\sigma = 1.5$} && \multicolumn{2}{c}{$\sigma = 2.0$}\\
	  \cmidrule{2-3} \cmidrule{5-6} \cmidrule{8-9} \cmidrule{11-12}
        & 
        \multicolumn{1}{c}{Error} & 
        \multicolumn{1}{c}{Time (s)} &&
        \multicolumn{1}{c}{Error} &
        \multicolumn{1}{c}{Time (s)} &&
        \multicolumn{1}{c}{Error} &
        \multicolumn{1}{c}{Time (s)} &&
        \multicolumn{1}{c}{Error} &
        \multicolumn{1}{c}{Time (s)}  \\
        \hline
BP &     0.0290 &     0.0020 &&     0.1683 &     0.0050 &&     0.1514 &     0.0023 &&     0.2809 &     0.0020 \\
B2-BP &     0.0217 &     0.0010 &&     0.1513 &     0.0037 &&     0.1414 &     0.0023 &&    \textbf{ 0.2520} &     0.0000 \\
B3-BP &     \textbf{0.0167} &     0.0000 &&     \textbf{0.1407 }&     0.0033 &&     \textbf{0.1394} &     0.0023 &&     0.2817 &     0.0000 \\
\hline
CBP-2 &     0.0073 &     0.0630 &&     0.0330 &     0.0827 &&     0.0712 &     0.0610 &&     0.2389 &     0.0467 \\
B2-CBP-2 &     0.0058 &     0.0500 &&     0.0272 &     0.0653 &&     0.0647 &     0.0487 &&     0.1977 &     0.0397 \\
B3-CBP-2 &     0.0056 &     0.0530 &&     0.0235 &     0.0670 &&     0.0469 &     0.0560 &&     0.1590 &     0.0450 \\
CBP-3 &     0.0052 &     0.1357 &&     0.0194 &     0.1653 &&     \textbf{0.0358} &     0.1313 &&     0.1074 &     0.1097 \\
B2-CBP-3 &     \textbf{0.0040} &     0.1093 &&     \textbf{0.0131} &     0.1347 &&     0.0368 &     0.1103 &&     \textbf{0.0919} &     0.0947 \\
\hline
TreeEP &     0.0101 &     0.0323 &&     0.0687 &     0.0473 &&     0.0815 &     0.0437 &&     0.0675 &     0.0487 \\
B2-TreeEP &     0.0100 &     0.0320 &&     0.0845 &     0.0553 &&     0.0878 &     0.0543 &&     0.0728 &     0.0457 \\
B3-TreeEP &     \textbf{0.0096} &     0.0350 &&     \textbf{0.0576} &     0.0587 && \textbf{0.0650} &  0.0907 && \textbf{0.0443} &     0.0780 \\
\hline
GBP-3 &     0.0290 &     1.1053 &&     0.1683 &     1.4173 &&     0.1514 &     1.0257 &&     0.2299 &     0.7763 \\
B2-GBP-3 &     \textbf{0.0217} &     0.8870 &&     \textbf{0.1513} &     1.2520 &&     \textbf{0.1414} &     0.9837 &&     \textbf{0.2280} &     0.6937 \\
GBP-4 &     0.0230 &     1.0157 &&     0.1548 &     1.4403 &&     0.1439 &     1.1250 &&     0.2286 &     0.7587 \\
    \bottomrule
    \end{tabular}
}}
\end{table}

\begin{table}
\centering
{\myfontsize{
\caption{Random Regular Graph with $p = 70$ Nodes, Degree $3$, and Attractive potentials: $20$ Trials}
\label{tab:randgraph70}
   \begin{tabular}{l r r c r r c r r c r r} \toprule
        \multirow{2}{*}{Algorithm} & \multicolumn{2}{c}{$\sigma = 0.5$} && \multicolumn{2}{c}{$\sigma = 1$} && 
	\multicolumn{2}{c}{$\sigma = 1.5$} && \multicolumn{2}{c}{$\sigma = 2.0$}\\
	  \cmidrule{2-3} \cmidrule{5-6} \cmidrule{8-9} \cmidrule{11-12}
        & 
        \multicolumn{1}{c}{Error} & 
        \multicolumn{1}{c}{Time (s)} &&
        \multicolumn{1}{c}{Error} &
        \multicolumn{1}{c}{Time (s)} &&
        \multicolumn{1}{c}{Error} &
        \multicolumn{1}{c}{Time (s)} &&
        \multicolumn{1}{c}{Error} &
        \multicolumn{1}{c}{Time (s)}  \\
        \hline
BP &     0.0172 &     0.0070 &&     0.1313 &     0.0200 &&     0.2144 &     0.0100 &&     0.2410 &     0.0100 \\
B2-BP &     0.0154 &     0.0040 &&     0.1211 &     0.0120 &&     0.2071 &     0.0090 &&     0.2895 &     0.0040 \\
B3-BP &    \textbf{ 0.0106} &     0.0025 &&     \textbf{0.0871} &     0.0120 &&    \textbf{ 0.1866} &     0.0065 &&     \textbf{0.2037} &     0.0040 \\
\hline
CBP-2 &     0.0069 &     0.0990 &&     0.0397 &     0.1425 &&     0.1036 &     0.1185 &&     0.1833 &     0.0965 \\
B2-CBP-2 &     0.0069 &     0.0870 &&     0.0459 &     0.1220 &&     0.0923 &     0.1080 &&     0.2571 &     0.0875 \\
B3-CBP-2 &     0.0063 &     0.0870 &&     0.0271 &     0.1135 &&     0.0882 &     0.1065 &&     0.1542 &     0.0880 \\
CBP-3 &     \textbf{0.0057} &     0.2120 &&     \textbf{0.0247} &     0.2840 &&     0.0811 &     0.2465 &&     0.1547 &     0.2050 \\
B2-CBP-3 &     0.0063 &     0.1875 &&     0.0264 &     0.2545 &&     \textbf{0.0727} &     0.2285 &&     \textbf{0.0897} &     0.1925 \\
\hline
TreeEP &     \textbf{0.0044} &     0.0500 &&     0.0404 &     0.0860 &&     0.1028 &     0.0985 &&     0.0741 &     0.1025 \\
B2-TreeEP &     0.0054 &     0.0575 &&     0.0484 &     0.1130 &&     0.1016 &     0.0990 &&     0.0969 &     0.1000 \\
B3-TreeEP &     0.0049 &     0.0715 &&     \textbf{0.0341} &     0.1005 &&     \textbf{0.0682} &     0.1575 &&     \textbf{0.0548} &     0.1730 \\
\hline
GBP-3 &     0.0150 &     1.4935 &&     0.1081 &     3.0275 &&     0.2100 &     2.2155 &&     0.2149 &     2.0235 \\
B2-GBP-3 &     \textbf{0.0127} &     1.4910 &&     \textbf{0.0939} &     2.8880 &&     \textbf{0.1998} &     2.1770 &&     \textbf{0.2046} &     1.6230 \\
    \bottomrule
    \end{tabular}
}}
\end{table}

\section{Summary}
\label{sec:summary}

We proposed a framework for generalizing inference (computing marginal distributions given the joint probability distribution) algorithms over graphical models (see Fig.~\ref{fig:generalizedAlg}).  The key components in our framework are (i) constructing a block-tree, a tree-structured graph over non-overlapping clusters, and (ii) constructing a block-graph, a graph over non-overlapping clusters.  We proposed a linear time algorithm for constructing block-trees and showed how large clusters in a block-tree can be split in a systematic manner to construct block-graphs that are favorable for inference.  Using numerical simulations, we showed that our framework for generalized inference in general leads to improved marginal estimates for many approximate inference algorithms implemented in the libDAI software package.  This suggests that the generalized inference framework can be used as a wrapper for improving the performance of approximate inference algorithms.  All the code and graphical models used in the numerical simulations can be downloaded from \url{http://www.ima.umn.edu/~dvats/GeneralizedInference.html}. Although the focus in this paper was on computing marginal estimates, our proposed block-graph based framework can also be used to generalize algorithms for computing the partition function ($Z$ in (\ref{eq:dist_mrf})) \cite{WainwrightTRW2005,ICML2011LiuIhler2011} or for the problem of MAP inference \cite{WainwrightMAP2005,KolmogorovCuts2006,SontagJaakkola2009,Jebara09mapestimation}.

There are several interesting research directions that can be further pursued to improve our generalized inference framework.  Our algorithm for constructing block-graphs only used the structure of the graph in computing the set of non-overlapping clusters.  Using the parameters of the graphical model may result in improved marginal estimates.  Further, it may be of interest to design block-graphs that are specific to the inference algorithm of interest.  Another interesting research direction is to combine frameworks that choose overlapping clusters with the block-graph framework.

\bibliographystyle{IEEEtran}
\bibliography{IEEEabrv,GeneralizedInference2012}

\begin{thebibliography}{10}
\providecommand{\url}[1]{#1}
\csname url@samestyle\endcsname
\providecommand{\newblock}{\relax}
\providecommand{\bibinfo}[2]{#2}
\providecommand{\BIBentrySTDinterwordspacing}{\spaceskip=0pt\relax}
\providecommand{\BIBentryALTinterwordstretchfactor}{4}
\providecommand{\BIBentryALTinterwordspacing}{\spaceskip=\fontdimen2\font plus
\BIBentryALTinterwordstretchfactor\fontdimen3\font minus
  \fontdimen4\font\relax}
\providecommand{\BIBforeignlanguage}[2]{{%
\expandafter\ifx\csname l@#1\endcsname\relax
\typeout{** WARNING: IEEEtran.bst: No hyphenation pattern has been}%
\typeout{** loaded for the language `#1'. Using the pattern for}%
\typeout{** the default language instead.}%
\else
\language=\csname l@#1\endcsname
\fi
#2}}
\providecommand{\BIBdecl}{\relax}
\BIBdecl

\bibitem{Reu2005}
H.~Rue and L.~Held, \emph{{G}aussian {M}arkov Random Fields: Theory and
  Applications (Monographs on Statistics and Applied Probability)},
  1st~ed.\hskip 1em plus 0.5em minus 0.4em\relax Chapman \& Hall/CRC, February
  2005.

\bibitem{WainwrightJordan2008}
M.~J. Wainwright and M.~I. Jordan, \emph{Graphical Models, Exponential
  Families, and Variational Inference}.\hskip 1em plus 0.5em minus 0.4em\relax
  Hanover, MA, USA: Now Publishers Inc., 2008.

\bibitem{KollerFriedman2009}
D.~Koller and N.~Friedman, \emph{Probabilistic Graphical Models: Principles and
  Techniques}.\hskip 1em plus 0.5em minus 0.4em\relax The MIT Press, 2009.

\bibitem{LauritzenSpiegelhalter1988}
S.~L. Lauritzen and D.~J. Spiegelhalter, ``Local computations with
  probabilities on graphical structures and their application to expert
  systems,'' \emph{Journal of the Royal Statistical Society. Series B
  (Methodological)}, vol.~50, no.~2, pp. 157--224, 1988.

\bibitem{tfabre-isit00}
E.~Fabre and A.~Guyader, ``Dealing with short cycles in graphical codes,'' in
  \emph{IEEE International Symposium on Information Theory (ISIT)}, June 2000,
  p.~10.

\bibitem{XingGMFPartition2004}
\BIBentryALTinterwordspacing
E.~P. Xing, M.~I. Jordan, and S.~Russell, ``Graph partition strategies for
  generalized mean field inference,'' in \emph{Proceedings of the 20th
  conference on Uncertainty in artificial intelligence}, ser. UAI '04.\hskip
  1em plus 0.5em minus 0.4em\relax Arlington, Virginia, United States: AUAI
  Press, 2004, pp. 602--610. [Online]. Available:
  \url{http://portal.acm.org/citation.cfm?id=1036843.1036916}
\BIBentrySTDinterwordspacing

\bibitem{Pearl1988}
J.~Pearl, \emph{Probabilistic Reasoning in Intelligent Systems: Networks of
  Plausible Inference}.\hskip 1em plus 0.5em minus 0.4em\relax Morgan Kaufmann,
  1988.

\bibitem{Eaton09}
F.~Eaton and Z.~Ghahramani, ``Choosing a variable to clamp: Approximate
  inference using conditioned belief propagation,'' in \emph{Proceedings of the
  Twelfth International Conference on Artificial Intelligence and Statistics},
  2009, pp. 145--152.

\bibitem{MontanariRizzo2005}
A.~Montanari and T.~Rizzo, ``How to compute loop corrections to the {B}ethe
  approximation,'' \emph{Journal of Statistical Mechanics: Theory and
  Experiment}, vol. 2005, no.~10, p. P10011, 2005.

\bibitem{MooijLC2007}
J.~M. Mooij and H.~J. Kappen, ``Loop corrections for approximate inference on
  factor graphs,'' \emph{J. Mach. Learn. Res.}, vol.~8, pp. 1113--1143, May
  2007.

\bibitem{MateescuKGDIJGP2010}
R.~Mateescu, K.~Kask, V.~Gogate, and R.~Dechter, ``Join-graph propagation
  algorithms,'' \emph{Journal of Artificial Intelligence Research}, vol.~37,
  pp. 279--328, 2010.

\bibitem{YedidaGBP2001}
J.~S. Yedidia, W.~T. Freeman, and Y.~Weiss, ``Generalized belief propagation,''
  in \emph{NIPS}, 2001, pp. 689--695.

\bibitem{HAK2003}
T.~Heskes, K.~Albers, and B.~Kappen, ``Approximate inference and constrained
  optimization,'' in \emph{UAI}, 2003, pp. 313--320.

\bibitem{YedidaTIT2005}
J.~S. Yedidia, W.~T. Freeman, and Y.~Weiss, ``Constructing free energy
  approximations and generalized belief propagation algorithms.'' \emph{IEEE
  Transactions on Information Theory}, vol.~51, no.~7, pp. 2282--2312, July
  2005.

\bibitem{Pelizzola2005}
A.~Pelizzola, ``Cluster variation method in statistical physics and
  probabilistic graphical models,'' \emph{Journal of Physics A: Mathematical
  and General}, vol.~38, no.~33, pp. R309--R339, 2005.

\bibitem{JordanVariational1999}
M.~Jordan, Z.~Ghahramani, T.~Jaakkola, and L.~Saul, ``An introduction to
  variational methods for graphical models,'' \emph{Machine Learning}, vol.~37,
  no.~2, pp. 183--233, Nov 1999.

\bibitem{YedidaTRBethe2001}
J.~Yedidia, W.~Freeman, and Y.~Weiss, ``{B}ethe free energy, {K}ikuchi
  approximations, and belief propagation algorithms,'' Mitsubishi Electric
  Research Laboratories, Tech. Rep. TR2001-16, 2001.

\bibitem{Kikuchi1951}
R.~Kikuchi, ``A theory of cooperative phenomena,'' \emph{Phys. Rev.}, vol.~81,
  no.~6, pp. 988-- 988--1003, 1951.

\bibitem{WellingUAI2004}
M.~Welling, ``On the choice of regions for generalized belief propagation,'' in
  \emph{Proceedings of the 20th conference on Uncertainty in Artificial
  Intelligence}, 2004, pp. 585--592.

\bibitem{WelMinTeh2005}
M.~Welling, T.~Minka, and Y.~W. Teh, ``Structured region graphs: Morphing {EP}
  into {GBP},'' in \emph{Proceedings of the International Conference on
  Uncertainty in Artificial Intelligence}, vol.~21, 2005.

\bibitem{XingGMF2003}
\BIBentryALTinterwordspacing
E.~P. Xing, M.~I. Jordan, and S.~Russell, ``A generalized mean field algorithm
  for variational inference in exponential families,'' in \emph{Proceedings of
  the 20th conference on Uncertainty in artificial intelligence}, ser. UAI
  '04.\hskip 1em plus 0.5em minus 0.4em\relax Arlington, Virginia, United
  States: AUAI Press, 2003, pp. 602--610. [Online]. Available:
  \url{http://portal.acm.org/citation.cfm?id=1036843.1036916}
\BIBentrySTDinterwordspacing

\bibitem{MarlinMurphy2009}
B.~M. Marlin and K.~P. Murphy, ``Sparse {G}aussian graphical models with
  unknown block structure,'' in \emph{International Conference on Machine
  Learning}, 2009, pp. 89--712.

\bibitem{JalaliRaviSang2011}
A.~Jalali, P.~Ravikumar, V.~Vasuki, and S.~Sanghavi, ``On learning discrete
  graphical models using group-sparse regularization,'' in \emph{International
  Conference on Machine Learning}, 2011, pp. 89--712.

\bibitem{MalioutovThesis}
D.~Malioutov, ``Approximate inference in {G}aussian graphical models,'' Ph.D.
  dissertation, Department of Electrical Engineering and Computer Science,
  Massachusetts Institute of Technology, 2008.

\bibitem{Levy1956}
P.~L{\'e}vy, ``A special problem of {B}rownian motion, and a general theory of
  {G}aussian random functions,'' in \emph{Proceedings of the {T}hird {B}erkeley
  {S}ymposium on {M}athematical {S}tatistics and {P}robability, 1954--1955,
  vol. {II}}.\hskip 1em plus 0.5em minus 0.4em\relax Berkeley and Los Angeles:
  University of California Press, 1956, pp. 133--175.

\bibitem{VatsMoura2009j}
D.~Vats and J.~M.~F. Moura, ``Telescoping recursive representations and
  estimation of {G}auss-{M}arkov random fields,'' \emph{Trans. on Information
  Theory}, vol.~57, no.~3, pp. 1645 -- 1663, 2011.

\bibitem{KalmanBucy1961}
R.~E. Kalman and R.~Bucy, ``New results in linear filtering and prediction
  theory,'' \emph{Transactions of the ASME--Journal of Basic Engineering},
  vol.~83, no. Series D, pp. 95--108, 1960.

\bibitem{RauchTungStriebel1965}
H.~E. Rauch, F.~Tung, and C.~T. Stribel, ``Maximum likelihood estimates of
  linear dynamical systems,'' \emph{AIAA J.}, vol.~3, no.~8, pp. 1445--1450,
  August 1965.

\bibitem{Lauritzen1996}
S.~L. Lauritzen, \emph{Graphical Models}.\hskip 1em plus 0.5em minus
  0.4em\relax Oxford University Press, USA, 1996.

\bibitem{Besag1974}
J.~Besag, ``Spatial interaction and the statistical analysis of lattice
  systems,'' \emph{Journal of the Royal Statistical Society. Series B
  (Methodological)}, vol.~36, no.~2, pp. 192--236, 1974.

\bibitem{HalinTreeDecomp1976}
R.~Halin, ``S-functions for graphs,'' \emph{Journal of Geometry}, vol.~8, pp.
  171--186, 1976, 10.1007/BF01917434.

\bibitem{Robertson1986}
N.~Robertson and P.~D. Seymour, ``Graph minors. {II}. {A}lgorithmic aspects of
  tree-width,'' \emph{Journal of Algorithms}, vol.~7, no.~3, pp. 309 -- 322,
  1986.

\bibitem{JensenJenson1994}
F.~Jensen and F.~Jensen, ``Optimal junction trees,'' in \emph{Proceedings of
  the 10th Annual Conference on Uncertainty in Artificial Intelligence
  (UAI-94)}.\hskip 1em plus 0.5em minus 0.4em\relax San Francisco, CA: Morgan
  Kaufmann, 1994, pp. 360--36.

\bibitem{Markowitz1957}
H.~M. Markowitz, ``The elimination form of the inverse and its application to
  linear programming,'' \emph{Management Science}, vol.~3, no.~3, pp. 255--269,
  1957.

\bibitem{BerryElimination2003}
A.~Berry, P.~Heggernes, and G.~Simonet, ``The minimum degree heuristic and the
  minimal triangulation process,'' in \emph{Graph-Theoretic Concepts in
  Computer Science}, ser. Lecture Notes in Computer Science, H.~Bodlaender,
  Ed.\hskip 1em plus 0.5em minus 0.4em\relax Springer Berlin / Heidelberg,
  2003, vol. 2880, pp. 58--70.

\bibitem{Kjaerulff1990}
U.~B. Kjaerulff, ``Triangulation of graphs - algorithms giving small total
  state space,'' Department of Mathematics and Computer Science, Aalborg
  University, Denmark, Tech. Rep. Research Report R-90-09, 1990.

\bibitem{Cooper1984}
G.~F. Cooper, ``Nestor: A computer-based medical diagnostic aid that integrates
  causal and probabilistic knowledge,'' Ph.D. dissertation, Department of
  Computer Science, Stanford University, 1984.

\bibitem{PengReggia1986}
Y.~Peng and J.~A. Reggia, ``Plausibility of diagnostic hypotheses,'' in
  \emph{National Conference on Artificial Intelligence (AAAI'86)}, 1986, pp.
  140--145.

\bibitem{Woods1977}
J.~W. Woods and C.~Radewan, ``Kalman filtering in two dimensions,''
  \emph{{IEEE} Trans. Inf. Theory}, vol.~23, no.~4, pp. 473--482, Jul 1977.

\bibitem{MouraBalram1992}
J.~M.~F. Moura and N.~Balram, ``Recursive structure of noncausal
  {G}auss-{M}arkov random fields,'' \emph{{IEEE} Trans. Inf. Theory}, vol.
  IT-38, no.~2, pp. 334--354, March 1992.

\bibitem{LevyAdamsWillsky1990}
B.~C. Levy, M.~B. Adams, and A.~S. Willsky, ``Solution and linear estimation of
  2-{D} nearest-neighbor models,'' \emph{Proc. {IEEE}}, vol.~78, no.~4, pp.
  627--641, Apr. 1990.

\bibitem{ShaferShenoy1990}
G.~Shafer and P.~P. Shenoy, ``Probability propagation,'' \emph{Annals of
  Mathematics and Artificial Intelligence}, no. 1-4, pp. 327--352, 1990.

\bibitem{ZhangPoole1994}
N.~Zhang and D.~Poole, ``A simple approach to {B}ayesian network
  computations,'' in \emph{Proceedings of the Tenth Canadian Conference on
  Artificial Intelligence}, 1994, pp. 171--178.

\bibitem{Dechter1999}
R.~Dechter, ``{Bucket elimination: A unifying framework for reasoning},''
  \emph{{Artificial Intelligence}}, vol. {113}, no. {1-2}, pp. {41--85}, {Sep}
  {1999}.

\bibitem{KaskDechter2001}
K.~Kask, R.~Dechter, J.~Larrosa, and F.~Cozman, ``Bucket-tree elimination for
  automated reasoning,'' University of California, Irvine, Tech. Rep. R92,
  2001.

\bibitem{kolar-song-xing-aoas09}
M.~Kolar, L.~Song, A.~Ahmed, and E.~P. Xing, ``Estimating time-varying
  networks,'' \emph{Annals of Applied Statistics}, 2009.

\bibitem{ZhouWassLaff2010}
S.~Zhou, J.~Lafferty, and L.~Wasserman, ``Time varying undirected graphs,''
  \emph{Machine Learning}, vol.~80, pp. 295--319, 2010,
  10.1007/s10994-010-5180-0.

\bibitem{FactorGraphFrey2001}
F.~Kschischang, B.~Frey, and H.-A. Loeliger, ``Factor graphs and the
  sum-product algorithm,'' \emph{IEEE Transactions on Information Theory},
  vol.~47, no.~2, pp. 498 --519, feb 2001.

\bibitem{Minka03tree}
T.~Minka and Y.~Qi, ``Tree-structured approximations by expectation
  propagation,'' in \emph{Proc. Neural Information Processing Systems Conf.
  (NIPS)}, 2003, p. 2003.

\bibitem{Mooij_libDAI_10}
J.~M. Mooij, ``lib{DAI}: A free and open source {C++} library for discrete
  approximate inference in graphical models,'' \emph{Journal of Machine
  Learning Research}, vol.~11, pp. 2169--2173, Aug. 2010.

\bibitem{ijgpsoftware}
V.~Gogate, ``Iterative joint graph propagation,''
  \url{http://www.hlt.utdallas.edu/~vgogate/ijgp.html}.

\bibitem{WainwrightTRW2005}
M.~J. Wainwright, T.~S. Jaakkola, and A.~S. Willsky, ``A new class of upper
  bounds on the log partition function,'' \emph{IEEE Trans. on Information
  Theory}, vol.~51, no.~7, pp. 2313 -- 2335, July 2005.

\bibitem{ICML2011LiuIhler2011}
Q.~Liu and A.~Ihler, ``Bounding the partition function using {H}older's
  inequality,'' in \emph{Proceedings of the 28th International Conference on
  Machine Learning}, L.~Getoor and T.~Scheffer, Eds., June 2011, pp. 849--856.

\bibitem{WainwrightMAP2005}
M.~J. Wainwright, T.~S. Jaakkola, and A.~S. Willsky, ``{MAP} estimation via
  agreement on (hyper)trees: Message-passing and linear programming
  approaches,'' \emph{{IEEE} Trans. Inf. Theory}, vol.~51, no.~11, pp.
  3697--3717, Nov. 2005.

\bibitem{KolmogorovCuts2006}
V.~Kolmogorov, ``Convergent tree-reweighted message passing for energy
  minimization,'' \emph{IEEE Transactions on Pattern Analysis and Machine
  Intelligence}, vol.~28, pp. 1568--1583, 2006.

\bibitem{SontagJaakkola2009}
D.~Sontag and T.~Jaakkola, ``Tree block coordinate descent for {MAP} in
  graphical models,'' \emph{Journal of Machine Learning Research}, vol.~5, pp.
  544--551, 2009.

\bibitem{Jebara09mapestimation}
T.~Jebara, ``{MAP} estimation, message passing, and perfect graphs,'' in
  \emph{Proceedings of the Twenty-Fifth Conference on Uncertainty in Artificial
  Intelligence}, ser. UAI '09, 2009, pp. 258--267.

\end{thebibliography}

\end{document}